\documentclass[journal]{IEEEtai}

\usepackage[colorlinks,urlcolor=blue,linkcolor=blue,citecolor=blue]{hyperref}

\usepackage{color,array}

\usepackage{graphicx}

\usepackage{graphicx}
\usepackage{amsfonts}
\usepackage{amsmath}
\usepackage{multirow}
\usepackage[table,xcdraw]{xcolor}

\makeatletter
\renewcommand{\@makefntext}[1]{\noindent\hspace{0pt}\textsuperscript{\@thefnmark} #1}
\makeatother

\newcommand{\squeezeup}{\vspace{-1mm}}
\begin{document}

\title{Efficient CNN Compression via Multi-method Low Rank Factorization and Feature Map Similarity}

% \author{First A. Author, \IEEEmembership{Fellow, IEEE}, Second B. Author, and Third C. Author, Jr., \IEEEmembership{Member, IEEE}
% \thanks{This paragraph of the first footnote will contain the date on which you submitted your paper for review. It will also contain support information, including sponsor and financial support acknowledgment. For example, ``This work was supported in part by the U.S. Department of Commerce under Grant BS123456.'' }
% \thanks{The next few paragraphs should contain the authors' current affiliations, including current address and e-mail. For example, F. A. Author is with the National Institute of Standards and Technology, Boulder, CO 80305 USA (e-mail: author@boulder.nist.gov).}
% \thanks{S. B. Author, Jr., was with Rice University, Houston, TX 77005 USA. He is now with the Department of Physics, Colorado State University, Fort Collins, CO 80523 USA (e-mail: author@lamar.colostate.edu).}
% \thanks{T. C. Author is with the Electrical Engineering Department, University of Colorado, Boulder, CO 80309 USA, on leave from the National Research Institute for Metals, Tsukuba, Japan (e-mail: author@nrim.go.jp).}
% \thanks{This paragraph will include the Associate Editor who handled your paper.}}

% \markboth{Journal of IEEE Transactions on Artificial Intelligence, Vol. 00, No. 0, Month 2020}
% {First A. Author \MakeLowercase{\textit{et al.}}: Bare Demo of IEEEtai.cls for IEEE Journals of IEEE Transactions on Artificial Intelligence}

\author{Milad Kokhazadeh, Georgios Keramidas, and Vasilios Kelefouras}

% \markboth{Journal of IEEE Transactions on Artificial Intelligence, Vol. 00, No. 0, Month 2020}
% {Kokhazadeh \MakeLowercase{\textit{et al.}}: Efficient CNN Compression via Multi-method Low Rank Factorization and Feature Map Similarity}

\maketitle

\begin{abstract}
Low-Rank Factorization (LRF) is a widely adopted technique for compressing deep neural networks (DNNs). However, it faces several challenges, including optimal rank selection, a vast design space, long fine-tuning times, and limited compatibility with different layer types and decomposition methods.
This paper presents an end-to-end Design Space Exploration (DSE) methodology and framework for compressing convolutional neural networks (CNNs) that addresses all these issues.
We introduce a novel rank selection strategy based on feature map similarity, which captures non-linear interactions between layer outputs more effectively than traditional weight-based approaches. Unlike prior works, our method uses a one-shot fine-tuning process, significantly reducing the overall fine-tuning time. The proposed framework is fully compatible with all types of convolutional (Conv) and fully connected (FC) layers.
To further improve compression, the framework integrates three different LRF techniques for Conv layers and three for FC layers, applying them selectively on a per-layer basis. We demonstrate that combining multiple LRF methods within a single model yields better compression results than using a single method uniformly across all layers.
Finally, we provide a comprehensive evaluation and comparison of the six LRF techniques, offering practical insights into their effectiveness across different scenarios.
The proposed work is integrated into TensorFlow 2.x, ensuring compatibility with widely used deep learning workflows. Experimental results on 14 CNN models across eight datasets demonstrate that the proposed methodology achieves substantial compression with minimal accuracy loss, outperforming several state-of-the-art techniques.
\end{abstract}

\begin{IEEEImpStatement}
Deep neural network (DNN) compression is a critical challenge in artificial intelligence (AI), especially for deploying models in resource-constrained environments. This article addresses the limitations of existing low-rank factorization (LRF) methods, such as suboptimal rank selection, long fine-tuning times, and limited applicability across different layers. A novel end-to-end compression framework is proposed, featuring a feature map similarity–based rank selection strategy, one-shot fine-tuning, and hybrid decomposition support. Unlike prior works, the framework applies different LRF methods per layer and supports six decomposition algorithms across convolutional and fully connected layers. Integrated into TensorFlow 2.x, it achieves superior compression and accuracy trade-offs across 14 CNN models and eight datasets, outperforming state-of-the-art techniques such as Variational Bayesian Matrix Factorization and filter-based pruning. This contribution enables scalable, architecture-agnostic compression and offers a practical tool for accelerating DNN deployment in mobile AI, embedded systems, and edge computing scenarios. It also provides insights that may guide future research on compression-aware design and model optimization.

% The impact statement should not exceeed 150 words. This section offers an example that is expanded to have only and just 150 words to demonstrate the point. Here is an example on how to write an appropriate impact statement: Chatbots are a popular technology in online interaction. They reduce the load on human support teams and offer continuous 24-7 support to customers. However, recent usability research has demonstrated that 30\% of customers are unhappy with current chatbots due to their poor conversational capabilities and inability to emotionally engage customers. The natural language algorithms we introduce in this paper overcame these limitations. With a significant increase in user satisfaction to 92\% after adopting our algorithms, the technology is ready to support users in a wide variety of applications including government front shops, automatic tellers, and the gaming industry. It could offer an alternative way of interaction for some physically disable users.
\end{IEEEImpStatement}

% \begin{IEEEkeywords}
% Enter key words or phrases in alphabetical order, separated by commas. For a list of suggested keywords, send a blank e-mail to \href{mailto:keywords@ieee.org}{\underline{keywords@ieee.org}} or visit \href{http://www.ieee.org/organizations/pubs/ani_prod/keywrd98.txt}{\underline{http://www.ieee.org/organizations/pubs/ani\_prod/keywrd98.txt}}
% \end{IEEEkeywords}


\section{Introduction}~\label{sec:introduction}

\IEEEPARstart{D}{espite} the recent success of Transformer-based models~\cite{transformer-success}, convolutional neural networks (CNNs) remain widely used in computer vision, including tasks such as
image classification, object detection, and semantic segmentation~\cite{cnn-applications}.
CNNs are computationally intensive and require substantial memory resources, which limits their deployment on resource-constrained platforms such as mobile devices, embedded systems, and edge devices~\cite{rcd-challenges, rcd-challenges2}.

\begin{figure}[tbp]
\centerline{\includegraphics[width=0.49\textwidth]{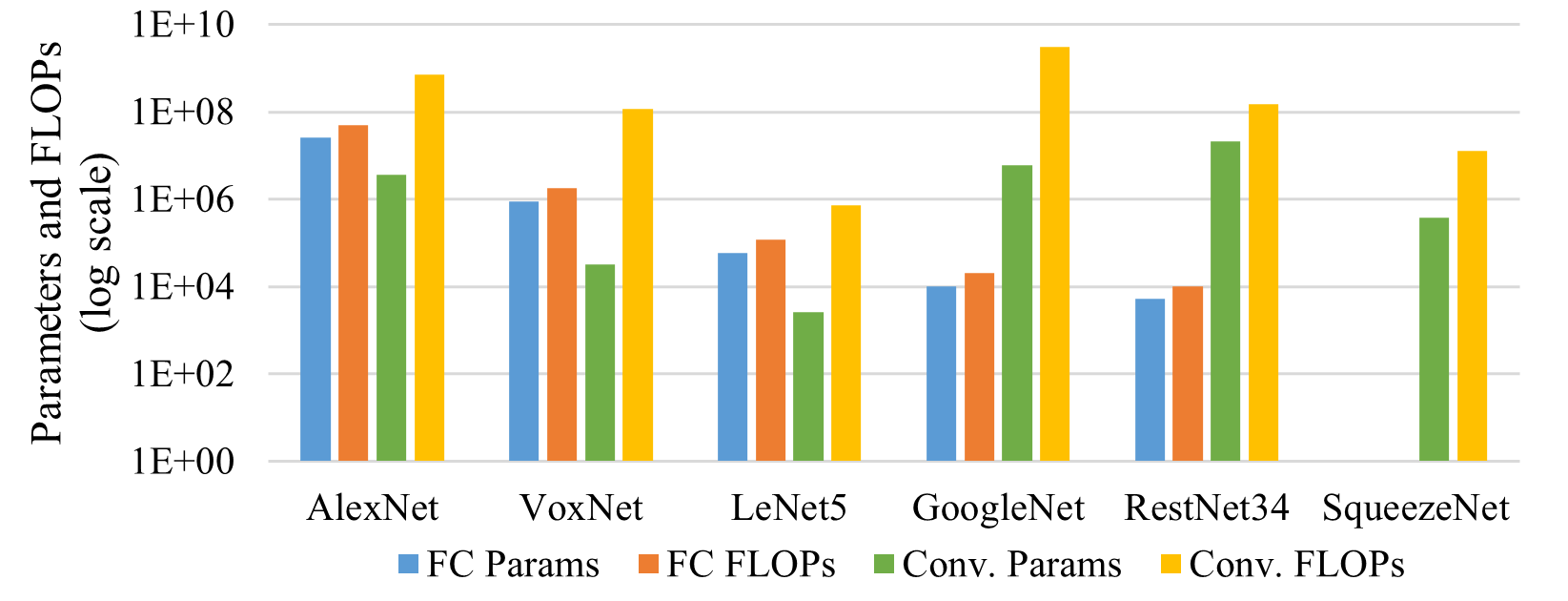}}
\caption{Breakdown of FLOPs and parameter distribution across Conv and FC layers in various CNN architectures.}

\squeezeup{}
\squeezeup{}
\squeezeup{}
\squeezeup{}
\squeezeup{}
\squeezeup{}

\label{fig:params_flops}
\end{figure}

Fig.~\ref{fig:params_flops} illustrates the number of parameters and Floating-Point Operations (FLOPs) associated with the Fully Connected (FC) and Convolutional (Conv) layers of various widely-used CNN models. As it is evident from Fig.~\ref{fig:params_flops}, CNN models exhibit diverse characteristics, emphasizing the need for a holistic compression methodology that effectively targets
both the FC and Conv layers within a unified framework.

Addressing this challenge has driven the development of numerous model compression techniques aimed at reducing model size and computational complexity while preserving accuracy~\cite{compression-techniques}.
Among the various model compression techniques, Low-Rank Factorization (LRF) has emerged as a particularly effective approach~\cite{tucker-2}.
LRF reduces both the number of parameters and the computational cost (FLOPs) of CNNs, making them more suitable for resource-constrained deployments.
Compared to other methods such as pruning~\cite{PR4}, quantization~\cite{quantization}, and knowledge distillation~\cite{knowledge-distillation}, LRF provides a flexible range of decomposition strategies, allowing for a fine-grained balance between memory usage, computational efficiency, and model accuracy~\cite{milad2}.

\begin{figure}[htbp]
\centerline{\includegraphics[width=0.45\textwidth]{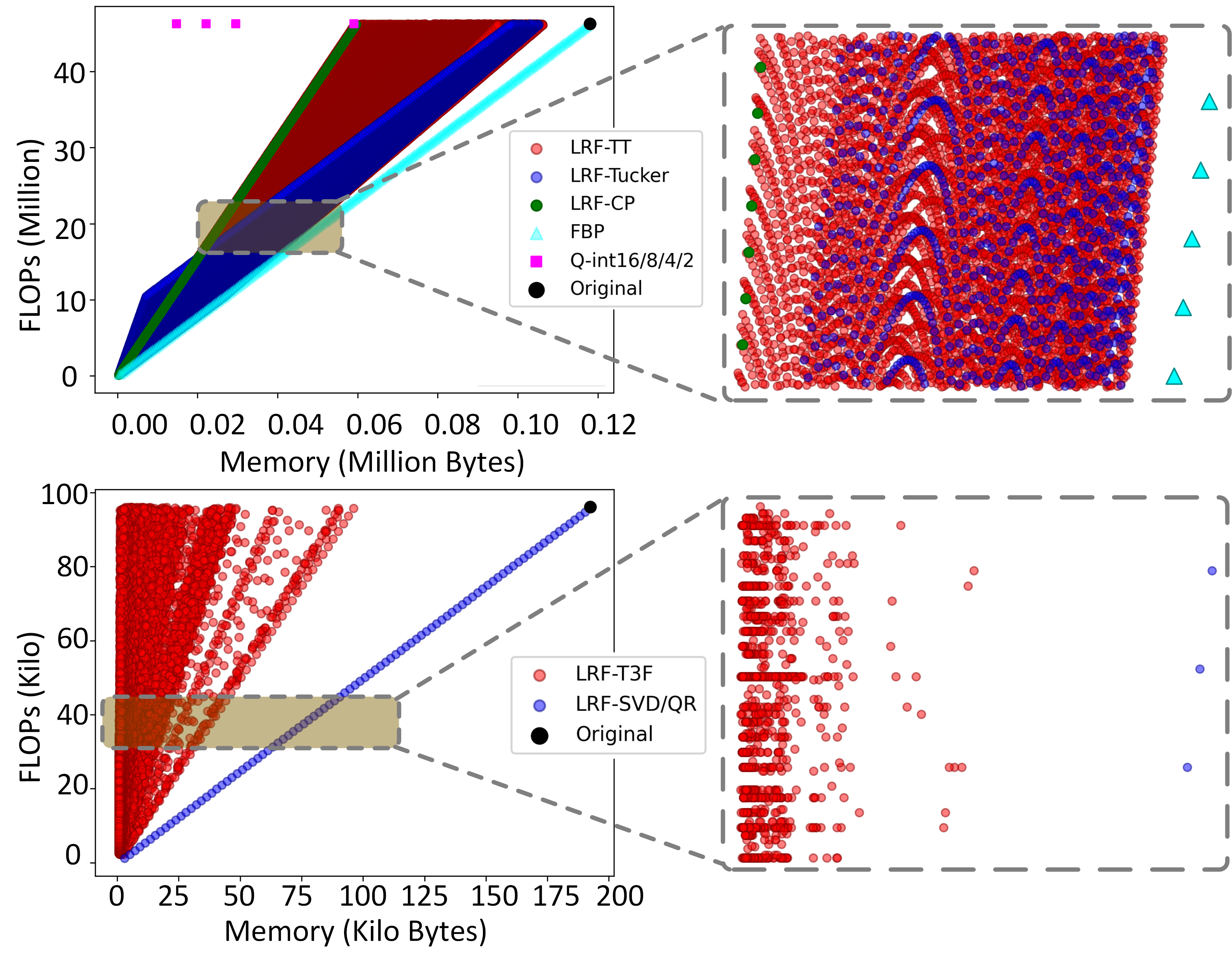}}

\squeezeup{}
\squeezeup{}
\squeezeup{}

\caption{Exploration space of multiple compression methods in the FLOPs-memory space. The graph at the top illustrates a Conv layer of ResNet50, while the graphs at the bottom shows a FC layer of LeNet5.}

\squeezeup{}
\squeezeup{}
\squeezeup{}

\label{fig:LRF_FBP_solutions}
\end{figure}

Setting aside evaluation details for now, the top graphs in Fig.~\ref{fig:LRF_FBP_solutions} compare model parameters and FLOPs for several DNN compression techniques applied to a Conv layer in ResNet50. Filter-based pruning (FBP) (cyan triangles) and quantization (pink squares), while somewhat effective, show limited ability to reduce both memory usage and computational cost simultaneously. For example, quantization reduces model size but does not decrease the number of parameters or FLOPs. FBP can reduce both parameters and FLOPs but falls short in achieving high compression ratios. In contrast, for the same FLOPs level, various LRF methods can achieve significantly greater compression.

However, the practical use of LRF in CNNs comes with significant challenges.
First, even a single layer presents a vast design space due to the many possible LRF configurations (Fig.~\ref{fig:LRF_FBP_solutions}). For example, applying Tucker decomposition to one layer of a ResNet model with shape (3, 3, 128, 256) results in 23,766 possible configurations. When extended to an entire model with multiple layers, the complexity becomes unmanageable for manual tuning.
Second, selecting the optimal rank for each layer typically requires costly iterative calibration or retraining.
For instance, in a relatively small model like LeNet-5, which has only 5 layers, the total number of possible LRF configurations can reach around 38 million.
Even under a modest evaluation setup-calibrating each configuration for just 5 epochs at 1 second per epoch—the total search time would exceed 2,193 days, underscoring the impracticality of exhaustive search for optimal settings.
Last but not least, existing LRF-based approaches often focus on specific layer types or decomposition methods, limiting their applicability to modern CNN architectures, which feature diverse layer types such as 1D, 2D, or 3D Conv layers, as well as FC layers.

This paper introduces an end-to-end Design Space Exploration (DSE) methodology and framework for CNN compression, addressing all the above challenges\footnote{This work is an extension of previous conference paper presented at the DATE 2025 conference~\cite{milad-date}}.
%Although the proposed approach is applicable to other DNNs and layers, its extension to these areas will be explored in future work.
The proposed methodology offers several key advantages/contributions over existing approaches:

\begin{itemize}
    \item \textbf{An end-to-end DSE framework} that formulates CNN compression using LRF as a \textbf{multi-objective} optimization problem balancing FLOPs, model parameters, overall memory usage, and validation accuracy.

    \item \textbf{Layer-wise, similarity-based rank selection:} We introduce an automated rank selection strategy that dynamically determines the optimal rank for each layer based on feature map similarity, rather than traditional weight similarity. This approach better captures the nonlinear relationships between features and enables more effective, tailored compression across layers.

    \item \textbf{Efficient fine-tuning process:} Our approach adopts a one-shot fine-tuning strategy that eliminates the need for iterative calibration or extensive retraining.

    \item \textbf{Compatibility with different layer types and decomposition methods:} The proposed framework supports a variety of layer types (e.g., 1D, 2D, and 3D Conv layers and FC layers) and multiple decomposition algorithms, i.e., Tucker decomposition~\cite{tucker}, Canonical Polyadic (CP) decomposition~\cite{cp}, Tensor Train (TT) decomposition~\cite{ttd}, Singular Value Decomposition (SVD)~\cite{svd}, QR~\cite{qr}, and T3F~\cite{t3f}).

    \item \textbf{Hybrid decomposition:} The proposed framework incorporates a post-processing step that combines three LRF methods for Conv layers (Tucker decomposition, CP decomposition, and TT decomposition) with three LRF methods for FC layers  (SVD, QR decomposition, and T3F). This hybrid approach achieves superior compression, enhancing overall model efficiency.

    \item \textbf{Compatibility with other compression techniques:} Our approach  is compatible with other compression techniques, like FBP and quantization, allowing seamless integration to achieve further reductions in model size.

    \item \textbf{A modular, easy-to-integrate framework built on TensorFlow 2.x~\cite{tensorflow}}, designed to plug seamlessly into existing deep learning workflows. The framework will be made publicly available upon acceptance to support reproducibility and further research.

    \item \textbf{An analytical study of six LRF methods}, offering insightful observations on their strengths and trade-offs.

\end{itemize}

We evaluate our approach on 14 popular CNNs across eight diverse datasets. Results show significant compression, averaging 77.8\%, 71.2\%, and 76\% for Tucker, CP, and TT (Conv layers), and 79.1\%, 79.7\%, and 81\% for SVD, QR, and T3F (FC layers), with under 1.5\% accuracy loss, outperforming several state-of-the-art techniques like VBMF ~\cite{VBMF}. Our hybrid approach achieves 82.5\% and 92.7\% average parameter reduction in Conv and FC layers, respectively.

The remainder of this paper is organized as follows. Section \ref{sec:background} provides the necessary background on LRF and CNN compression techniques, while Section \ref{sec:related-work} reviews related literature.
Section \ref{subsec:different-methods} evaluates the six studied methods, while Section \ref{sec:methodology} provides the proposed methodology and hybrid decomposition approach. Section \ref{sec:experimental-setup} describes the evaluation framework, and Section \ref{sec:results} presents the experimental results. Finally, Section \ref{sec:conclusion} concludes the paper.

\section{Background}~\label{sec:background}

LRF is a fundamental technique for compressing DNNs by exploiting the redundancy in their weight matrices and tensors. By approximating high-dimensional tensors or matrices with smaller, factorized components, these methods reduce both the number of parameters and FLOPs. Broadly, decomposition methods can be categorized into matrix and tensor decomposition methods.

\subsection{Matrix Decomposition Methods}~\label{subsec:background-fc}

\begin{figure}[tbp]
\centerline{\includegraphics[width=0.35\textwidth]{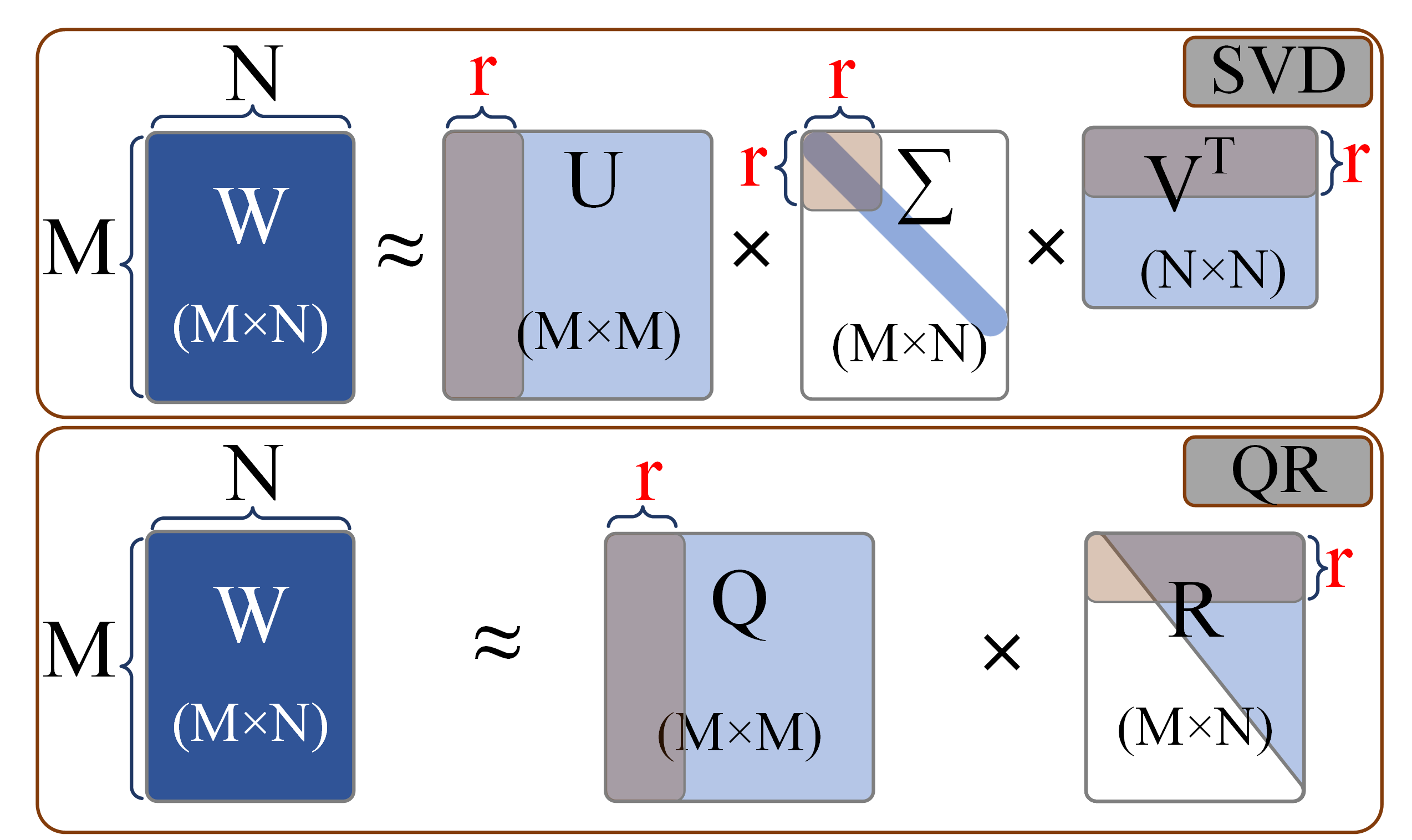}}

\squeezeup{}
\squeezeup{}
\squeezeup{}

\caption{Matrix decomposition using SVD (top) and QR decomposition (bottom)}

\label{fig:svd-qr}
\end{figure}

\noindent\textbf{SVD:}
The SVD method~\cite{svd} is depicted in the top of Fig.~\ref{fig:svd-qr}. It factorizes a given matrix W into three matrices whose size is smaller than the original matrix for a given rank $r$. When these matrices are multiplied together, they reconstruct the original matrix W.

\noindent\textbf{QR Decomposition:}
The QR decomposition method~\cite{qr}, shown in the bottom of Fig.~\ref{fig:svd-qr}, factors a matrix W into two components: an orthogonal matrix Q and an upper triangular matrix R.

\subsection{Tensor Decomposition Methods} \label{subsec:background-tensor}

\begin{figure}[tbp]
\centerline{\includegraphics[width=0.35\textwidth]{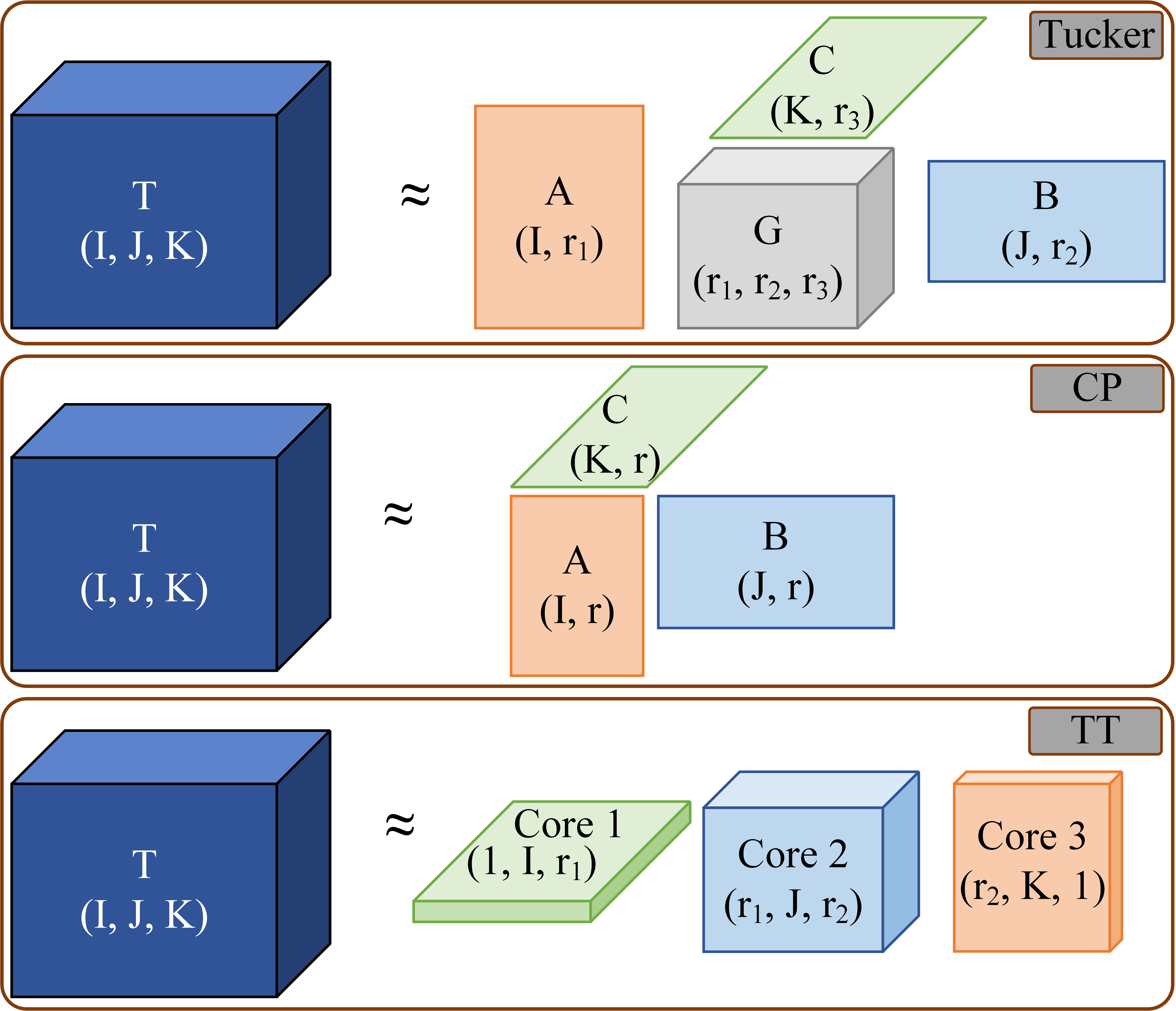}}

\squeezeup{}
\squeezeup{}
\squeezeup{}

\caption{Tensor decomposition of Conv2D layers using Tucker (top), CP (middle), and TT (bottom) methods.}

\squeezeup{}
\squeezeup{}
\squeezeup{}

\label{fig:tucker-cp-tt}
\end{figure}

\noindent \textbf{Tucker Decomposition:}
Tucker decomposition~\cite{tucker} is depicted at the top of Fig.~\ref{fig:tucker-cp-tt}. It decomposes a high-dimensional tensor $X$ into a smaller core tensor G, along with multiple factor matrices (A, B, and C as shown at the top of Fig.~\ref{fig:tucker-cp-tt}).

\noindent \textbf{CP Decomposition:}
CP decomposition~\cite{cp}, illustrated at the middle of Fig.~\ref{fig:tucker-cp-tt}, represents a high-order tensor as a sum of \textit{rank-1} tensors, effectively factorizing it into a set of vectors along each mode. Indeed, as shown in the middle of Fig.~\ref{fig:tucker-cp-tt} this results in a sequence of 2D matrices.

\noindent \textbf{TT Decomposition and T3F:}
TT decomposition~\cite{ttd}, shown at the bottom of Fig.~\ref{fig:tucker-cp-tt}, represents a high-dimensional tensor as a sequence of smaller, lower-dimensional tensors, commonly referred to as "cores," which are interconnected by contracted indices\footnote{For better visualization a 3D tensor is used in these cases}.
T3F ~\cite{t3f} is a library for employing TT decomposition to FC layers. T3F factorizes a FC layer by reshaping its 2D weight matrix into a higher-dimensional tensor. It then uses TT decomposition and Kronecker product to factorize a FC layer~\cite{Novikov2015}\footnote{Since the resulting factors are 4D tensors in the T3F, it is difficult to visualize it in a 2D figure, and thus we do not present them graphically.}.

\subsection{Decomposing Fully Connected (FC) layers}

Applying matrix decomposition methods to FC layers is straightforward since the weight matrix is inherently a 2D array. However, tensor decomposition methods, such as TT decomposition, can also be applied by reshaping the weight matrix of a FC layer into a tensor~\cite{t3f}, but this introduces an overhead.

\subsection{Decomposing Convolution (Conv) layers}

Conv layers naturally operate on tensors, making tensor decomposition methods more suitable and straightforward to apply. Conversely, matrix decomposition methods are less commonly used in Conv layers due to their 2D structure and lack of alignment with the multi-dimensional nature of Conv operations~\cite{joint-matrix}. However, for $1 \times 1$ Conv layers, the weight tensor reduces to a 2D array, allowing matrix decomposition methods to be applied directly.

\section{Related Work} \label{sec:related-work}

The application of LRF in CNNs has been extensively studied in the literature ~\cite{LRF3, LRF4, manual1, manual2}, with research efforts primarily addressing three key challenges: rank selection and large exploration space (ES), the lengthy fine-tuning/re-training process, and compatibility issues.
Specifically, selecting the appropriate ranks is critical for balancing compression and accuracy but since this is an NP-complete problem several studies resort to manual rank selection~\cite{LRF3, LRF4, manual1, manual2}.
The large ES introduces complexity, making it difficult to efficiently explore all possible solutions.
The fine-tuning or re-training process is often time-consuming, limiting the practicality of LRF in real-world applications.
Finally, compatibility remains a challenge, as no single LRF method can be universally applied to all CNN layers.

To address the issue of manual rank selection, various automated rank selection approaches have been developed, including reinforcement learning~\cite{reinforcement-learning} and genetic algorithms~\cite{genetic}.
However, nearly all of them encounter a similar issue: as the compression ratio increases, the time required also rises. This is due to the non-linear increase in complexity involved in finding the optimal LRF combination~\cite{LRF12}.

Several papers utilize analytical methods like Variational Bayesian Matrix Factorization (VBMF)~\cite{VBMF} and machine learning-based global optimization techniques, such as Bayesian Optimization~\cite{bayesopt}, to automatically select the rank of each layer~\cite{VBMF1, VBMF2}. These techniques focus on the weight tensor of individual layers, without taking into consideration: i) the influence of subsequent layers' outputs, e.g., activation, pooling, or batch normalization layers and ii) the effects of interactions among the multiple layers within the model.

To address the aforementioned problems, while some studies attempt to train a low-rank network from scratch~\cite{training1, training2}, others address the optimal rank selection problem by defining LRF within the context of network architecture search (NAS) process~\cite{NAS}. Nevertheless, all these methods require retraining the model, which is time-intensive. The works~\cite{milad1, milad2, milad3} address the challenge of large search space and rank selection in DNNs through a DSE methodology. However, their solutions are limited to FC layers and apply the same compression ratio across different FC layers, which may not fully optimize the compression and performance trade-offs for diverse network architectures.

\begin{figure}[t]
\centerline{\includegraphics[width=0.35\textwidth]{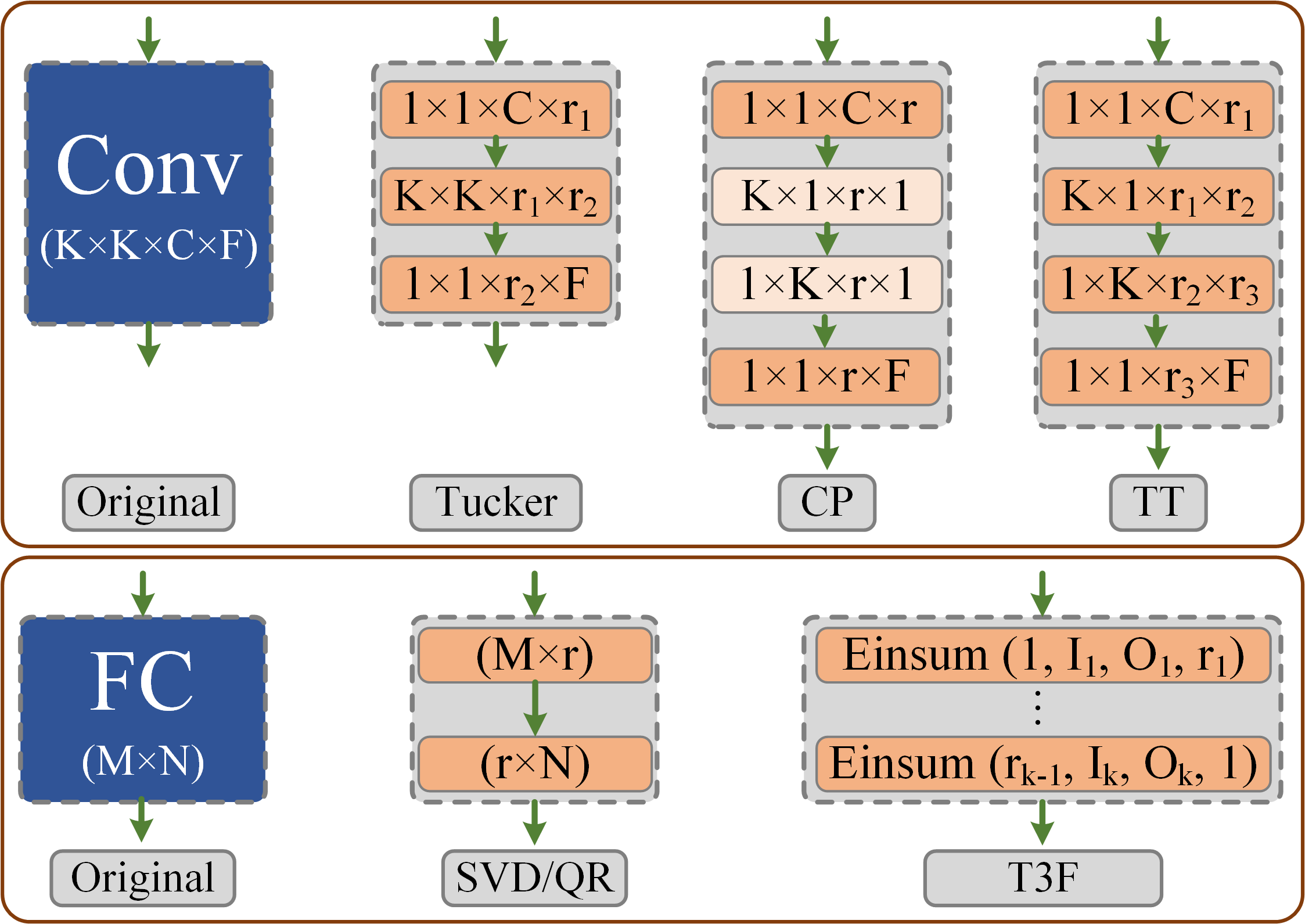}}

\squeezeup{}
\squeezeup{}
\squeezeup{}

\caption{LRF process for Conv2D layers using Tucker, CP, and TT decomposition (top), and for FC layers using SVD, QR, and T3F (bottom)}

\squeezeup{}
\squeezeup{}
\squeezeup{}

\label{fig:conv-fc-factorized}
\end{figure}

\begin{figure}[b]

\squeezeup{}
\squeezeup{}
\squeezeup{}

\centerline{\includegraphics[width=0.49\textwidth]{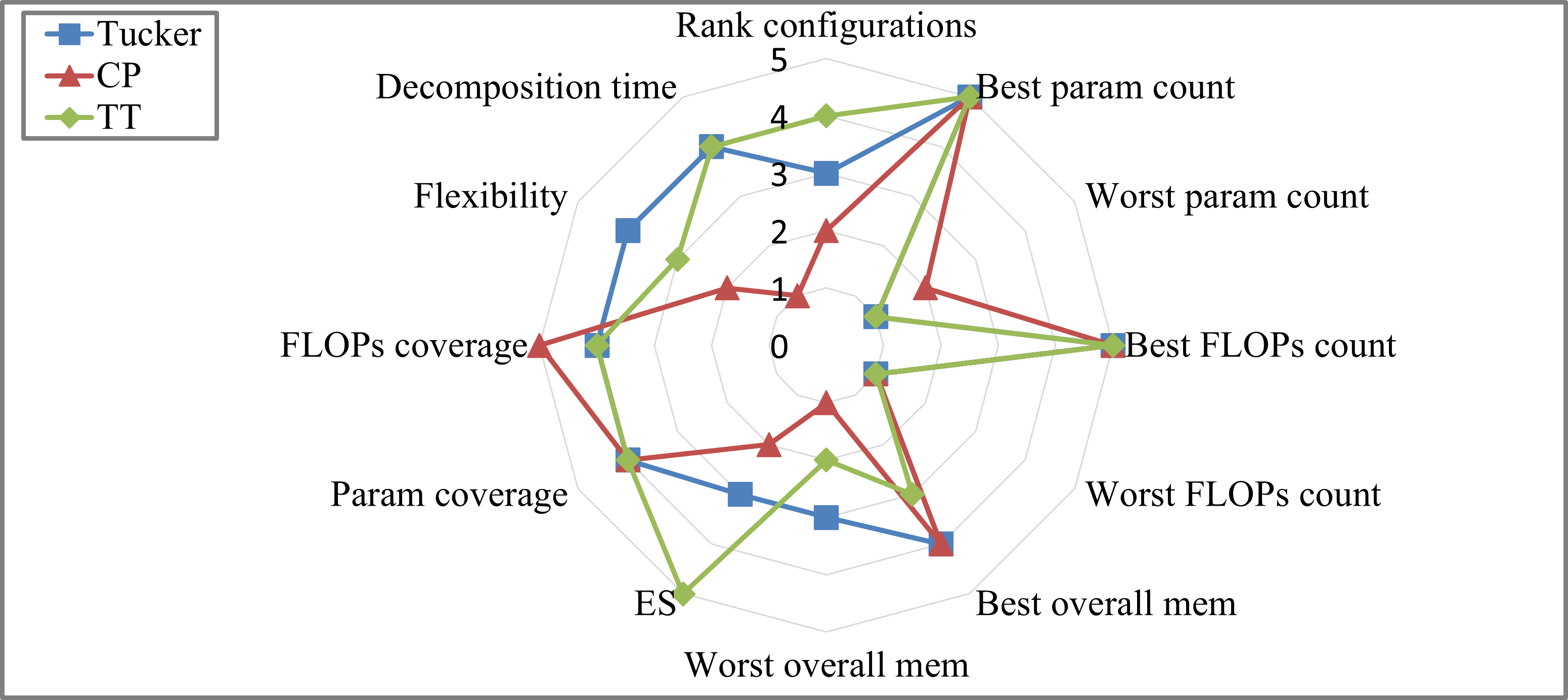}}

\squeezeup{}
\squeezeup{}
\squeezeup{}

\caption{Evaluation of the three LRF methods used for Conv layers (higher score means better).}
\label{fig:spider_conv}
\end{figure}

A closely related study to the proposed methodology is presented in~\cite{feature-map-norm}, where feature maps and their norms are utilized to estimate the rank of each layer. However, this approach treats each layer independently, without considering inter-layer interactions. Additionally, unlike our proposed one-shot fine-tuning strategy, their method employs a slower iterative process to refine the rank selection for each layer.

The first systematic comparison of decomposition methods is~\cite{informative-approximation-error}, which selects layer-wise ranks based on tensor approximation error. However, it lacks automatic rank selection, uses fixed compression ratios per method, treats methods in isolation, and does not offer a unified compression framework.

To enhance accuracy retention in low-rank Vision Transformers, an iterative, greedy selection metric is employed, incorporating cosine similarity to determine the rank of each layer \cite{LRF13}.
However, this approach is limited to the self-attention components of a specific Transformer model and focuses solely on the weights of each layer.

In the previous work~\cite{milad-date}, a one-shot and self-adaptive rank selection technique is prposed. This work extends and enhances it by integrating six LRF methods into the framework, analyzing the impact of key parameters, providing an analytical comparison of the LRF methods, and introducing a hybrid compression strategy that combines six decompositions within a single model.

\section{Analytical Study of LRF Techniques as a Basis for the Proposed Framework} \label{subsec:different-methods}

\begin{figure}[b]

\squeezeup{}
\squeezeup{}
\squeezeup{}

\centerline{\includegraphics[width=0.49\textwidth]{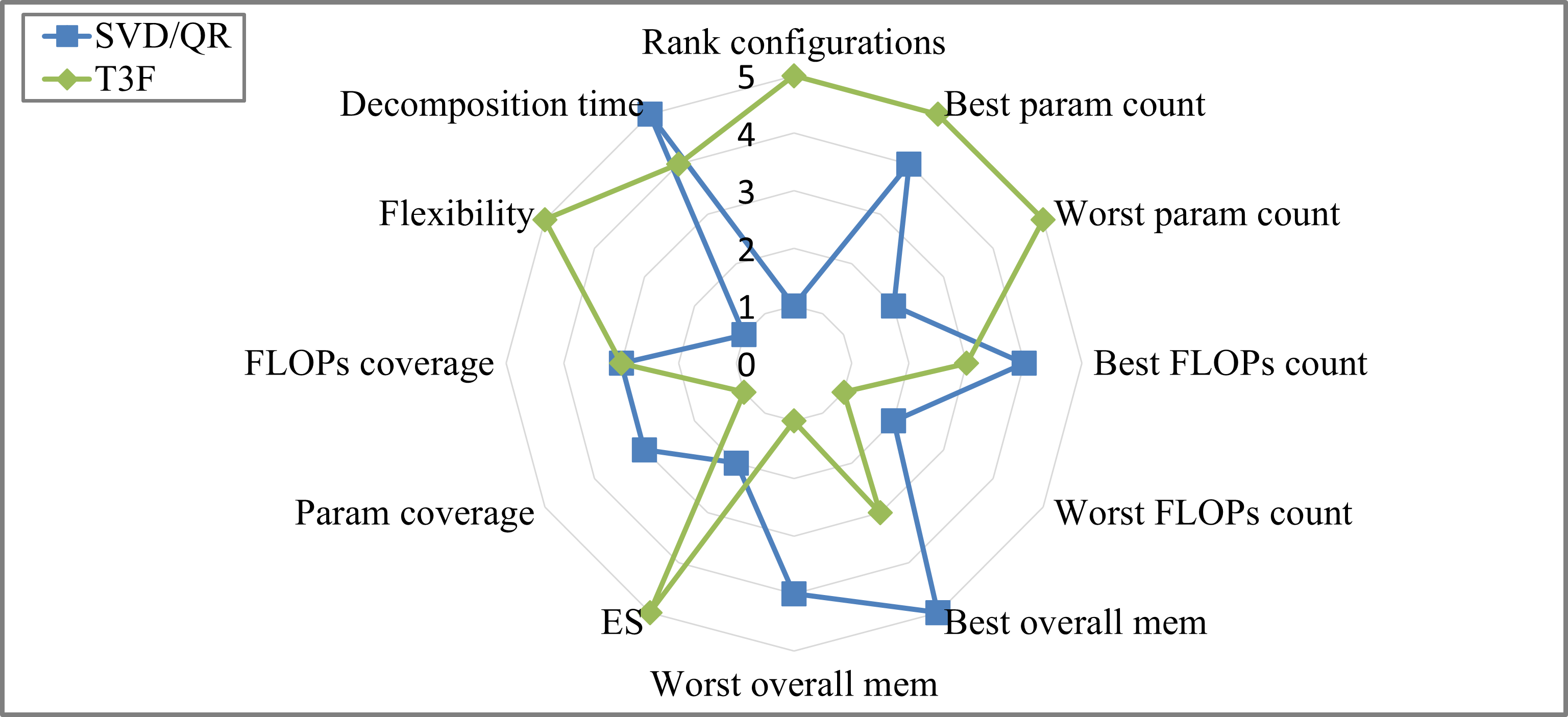}}

\squeezeup{}
\squeezeup{}
\squeezeup{}

\caption{Evaluation of the three LRF methods used for FC layers (higher score means better).}
\label{fig:spider_fc}
\end{figure}

In this Subsection, we conduct a thorough  analysis of the six LRF methods studied. This analysis is essential for developing the proposed framework, as it involves three key steps: first, deriving the mathematical equations for parameter count, overall memory, FLOPs, and the exploration space (ES) of each method; second, identifying the unique characteristics of each technique, such as which dimensions to factorize and the corresponding rank configurations; and third, establishing the foundation for the hybrid decomposition step.
We also provide insightful observations on the characteristics and strengths of the six, studied decomposition methods and deduct specific conclusions regarding their suitability for various scenarios.

\noindent \textbf{Rank Configuration.}
To fully understand the decomposition process, we first analyze which dimensions can be factorized and how many ranks can be used in each method. Fig.~\ref{fig:conv-fc-factorized} illustrates how an uncompressed layer is decomposed into multiple 'smaller' layers for Conv (top) and FC (bottom) layers, respectively.

For Conv layers, CP and TT factorize all tensor dimensions, while Tucker decomposition allows selective factorization across different dimensions, enabling decomposition of one, two, or more dimensions as needed. In this paper, we apply Tucker only to the input and output channel dimensions, leaving the spatial (kernel) dimensions intact, as kernel sizes are typically small and offer limited compressibility. For example, given a layer of shape (3, 3, 128, 256), we decompose the dimensions of size 128 and 256, corresponding to the input channels and filters, respectively.

% \usepackage{multirow}
\begin{table*}[h]
\caption{Evaluation in terms of Parameters (P), Feature Maps (FM), FLOPs (F), and Complexity (O).}

\centering

\begin{tabular}{
>{\columncolor[HTML]{DAE8FC}}c
>{\columncolor[HTML]{DAE8FC}}c l|
>{\columncolor[HTML]{DAE8FC}}c
>{\columncolor[HTML]{DAE8FC}}c l}
\hline
\multicolumn{3}{c|}{\cellcolor[HTML]{DAE8FC}\textbf{Convolution layer}} & \multicolumn{3}{c}{\cellcolor[HTML]{DAE8FC}\textbf{FC layer}} \\ \hline
\textbf{Method} & \multicolumn{2}{c|}{\cellcolor[HTML]{DAE8FC}\textbf{Equation}} & \textbf{Method} & \multicolumn{2}{c}{\cellcolor[HTML]{DAE8FC}\textbf{Equation}} \\ \hline
\cellcolor[HTML]{DAE8FC} & \textbf{P} & $K_1 \times K_2 \times C \times F$ & \cellcolor[HTML]{DAE8FC} & \textbf{P} & $M \times N$ \\ \cline{2-3} \cline{5-6}
\cellcolor[HTML]{DAE8FC} & \textbf{FM} & $X' \times Y' \times F$ & \cellcolor[HTML]{DAE8FC} & \textbf{FM} & $M$ \\ \cline{2-3} \cline{5-6}
\cellcolor[HTML]{DAE8FC} & \textbf{F} & $2 \times X' \times Y' \times F \times K_1 \times K_2 \times C$ & \cellcolor[HTML]{DAE8FC} & \textbf{F} & $2 \times M \times N$ \\ \cline{2-3} \cline{5-6}
\multirow{-4}{*}{\cellcolor[HTML]{DAE8FC}\textbf{Original}} & \textbf{O} & $O(X'Y' FCK_1K_2)$ & \multirow{-4}{*}{\cellcolor[HTML]{DAE8FC}\textbf{Original}} & \textbf{O} & $O(MN)$ \\ \hline
\cellcolor[HTML]{DAE8FC} & \textbf{P} & $(C \times r_1) + (K_1 \times K_2 \times r_1 \times r_2) + (r_2 \times F)$ & \cellcolor[HTML]{DAE8FC} & \textbf{P} & $(M \times r) + (r \times N)$ \\ \cline{2-3} \cline{5-6}
\cellcolor[HTML]{DAE8FC} & \textbf{FM} & $(X \times Y \times r_1)+(X' \times Y' \times r_2)+(X' \times Y' \times F)$ & \cellcolor[HTML]{DAE8FC} & \textbf{FM} & $r+M$ \\ \cline{2-3} \cline{5-6}
\cellcolor[HTML]{DAE8FC} & \textbf{F} & \begin{tabular}[c]{@{}l@{}}$2 \times [(X \times Y \times r_1 \times C)+ $\\ $(X' \times Y' \times r_2 \times K_1 \times K_2 \times r_1)+$\\ $(X' \times Y' \times F \times r_2)]$\end{tabular} & \cellcolor[HTML]{DAE8FC} & \textbf{F} & $2 \times [(M \times r)+ (r \times N)]$ \\ \cline{2-3} \cline{5-6}
\multirow{-4}{*}{\cellcolor[HTML]{DAE8FC}\textbf{Tucker}} & \textbf{O} & $O(X Y Cr_1+X'Y'r_1r_2K_1K_2+X'Y'Fr_2)$ & \multirow{-4}{*}{\cellcolor[HTML]{DAE8FC}\textbf{SVD}} & \textbf{O} & $O(r(M+N)$ \\ \hline
\cellcolor[HTML]{DAE8FC} & \textbf{P} & $(C \times r)+(K_1 \times r)+(K_2 \times r)+(r \times F)$ & \cellcolor[HTML]{DAE8FC} & \textbf{P} & $(M \times r) + (r \times N)$ \\ \cline{2-3} \cline{5-6}
\cellcolor[HTML]{DAE8FC} & \textbf{FM} & \begin{tabular}[c]{@{}l@{}}$(X \times Y \times r)+(X' \times Y \times r)+$\\ $(X' \times Y' \times r)+(X' \times Y' \times F)$\end{tabular} & \cellcolor[HTML]{DAE8FC} & \textbf{FM} & $r+M$ \\ \cline{2-3} \cline{5-6}
\cellcolor[HTML]{DAE8FC} & \textbf{F} & \begin{tabular}[c]{@{}l@{}}$2 \times [(X \times Y \times r \times C)+$ \\ $(X' \times Y  \times K_1 \times r)+$\\ $(X' \times Y' \times K_2 \times r)+$\\ $(X' \times Y' \times F \times r)]$\end{tabular} & \cellcolor[HTML]{DAE8FC} & \textbf{F} & $2 \times [(M \times r)+ (r \times N)]$ \\ \cline{2-3} \cline{5-6}
\multirow{-4}{*}{\cellcolor[HTML]{DAE8FC}\textbf{CP}} & \textbf{O} & $O(XYCr+X'YrK_1+X'Y'rK_2+X'Y'Fr)$ & \multirow{-4}{*}{\cellcolor[HTML]{DAE8FC}\textbf{QR}} & \textbf{O} & $O(r(M+N)$ \\ \hline
\cellcolor[HTML]{DAE8FC} & \textbf{P} & $(C \times r_1)+(K_1 \times r_1 \times r_2)+(K_2 \times r_2 \times r_3)+(r_3 \times F)$ & \cellcolor[HTML]{DAE8FC} & \textbf{P} & $\sum_{t=1}^d (r_{t-1}  \cdot m_t \cdot n_t \cdot r_t)$ \\ \cline{2-3} \cline{5-6}
\cellcolor[HTML]{DAE8FC} & \textbf{FM} & \begin{tabular}[c]{@{}l@{}}$(X \times Y \times r_1)+(X' \times Y \times r_2)+$\\ $(X' \times Y' \times r_3)+(X' \times Y' \times F)$\end{tabular} & \cellcolor[HTML]{DAE8FC} & \textbf{FM} & $\sum_{t=d}^{1}  m_t \cdot r_{t-1} \cdot (X_{t+1}/(n_d \cdot r_d))$ \\ \cline{2-3} \cline{5-6}
\cellcolor[HTML]{DAE8FC} & \textbf{F} & \begin{tabular}[c]{@{}l@{}}$2 \times [(X \times Y \times r_1 \times C)+ $\\ $(X' \times Y \times r_2 \times K_1 \times r_1)+$\\ $(X' \times Y' \times r_3 \times K_2 \times r_2)+$\\ $(X' \times Y' \times F \times r_3)]$\end{tabular} & \cellcolor[HTML]{DAE8FC} & \textbf{F} & $\sum_{t=1}^{d} 2\cdot r_t \cdot r_{t-1} \cdot m_{t}\cdot \ldots m_d\cdot n_1\cdot \ldots \cdot n_{t}$ \\ \cline{2-3} \cline{5-6}
\multirow{-4}{*}{\cellcolor[HTML]{DAE8FC}\textbf{TT}} & \textbf{O} & $O(XYCr_1+X'Yr_1r_2K_1+X'Y'r_2r_3K_2+X'Y'Fr_3)$ & \multirow{-4}{*}{\cellcolor[HTML]{DAE8FC}\textbf{T3F}} & \textbf{O} & $O(d max\{r_i\}^2 max\{n_i\} max\{M, N\})$ \\ \hline
\end{tabular}
\label{table:equations}
\end{table*}

CP employs a single rank value, whereas TT uses five rank values where the first and last ranks are always equal to 1.
Regarding Tucker, we choose to use two rank values in this paper.
As we will discuss later, using multiple ranks enhances flexibility by enabling independent control over the factorization along different tensor dimensions.
This is important because the redundancy in weight tensors is often unevenly distributed, e.g., some dimensions (e.g., output channels) may be more compressible than others (e.g., spatial dimensions). By assigning separate rank values per dimension, the decomposition can better exploit this anisotropy, leading to more effective compression without uniformly degrading performance.

In the case of FC layers, SVD and QR factorize both dimensions and employ a single rank value (Fig.~\ref{fig:conv-fc-factorized}). In contrast, TT, as implemented in the T3F library~\cite{t3f} for FC layers, requires a list of ranks along with a predefined tensor shape to which the original 2D matrix is transformed. Notably, similar to standard TT decomposition, the first and last ranks in T3F are always constrained to one.

\noindent \textbf{Parameter Count.}
Table~\ref{table:equations} summarizes the mathematical expressions for calculating the number of parameters (P), FLOPs (F), feature map (FM) elements, and computational complexity (Big O notation), for each decomposition method.

To compare the six LRF methods in terms of memory and FLOPs, we adopt two complementary approaches:
\begin{enumerate}
  \item Max/Min Analysis: We compute the maximum and minimum achievable parameter count, overall memory and FLOPs for each method (Fig.~\ref{fig:spider_conv} and Fig.~\ref{fig:spider_fc}).
  \item Constraint-based Analysis: We evaluate them under specific memory and FLOPs constraints. Specifically, which method achieves the lowest FLOPs for a given level of memory compression, and which method achieves the best parameter count for a fixed reduction in FLOPs.
\end{enumerate}

Fig.~\ref{fig:spider_conv} and Fig.~\ref{fig:spider_fc} present the average values across all layers from 14 CNN architectures studied in this work (detailed in the Appendix).
Note that the memory and FLOPs equations for SVD and QR are identical (Table~\ref{table:equations}); therefore, a single line (blue) represents both methods in Fig.~\ref{fig:spider_fc}.
It is important to note that this analysis does not take into account the model's accuracy. To the best of our knowledge, this is the first work that provides a quantitative comparison of various LRF methods.

Regarding the first way of evaluation and Conv layers, all methods exhibit similar best achievable parameter count (with a maximum difference of only 1\%). However, their worst-case compression varies significantly (Fig.~\ref{fig:spider_conv} and Fig.~\ref{fig:spider_fc}).For FC layers, T3F achieves the highest parameter reduction, benefiting from its ability to exploit multi-dimensional redundancies more effectively than the matrix decomposition methods.

In the second evaluation method (not shown in Fig.\ref{fig:spider_conv} and Fig.\ref{fig:spider_fc}), LRF methods yield significantly different results under fixed memory or FLOPs constraints, depending on the layer’s shape. For example, for a layer of shape (3, 3, 256, 512) and 60\% parameter reduction, Tucker offers two solutions with 47\% and 53\% FLOPs reduction, CP gives one with 20\%, and TT provides 141 solutions ranging from 1\% to 57\%. Similarly, fixing FLOPs can lead to large variations in parameter count across methods.

\noindent \textbf{Overall Memory.}
The overall memory footprint consists of the memory required for both the  parameters and the FMs.

In Conv layers, FMs have a minor impact on overall memory when spatial dimensions are small but can dominate memory usage when spatial dimensions are large, often exceeding the memory required for the parameters.
Moreover, at low compression levels, TT and CP can produce FMs larger than those of the original layer.
In FC layers, SVD and QR decompositions generate minimal FM memory, making their contribution to the overall footprint negligible. In contrast, T3F produces multiple high-memory FMs, significantly increasing overall memory.

Regarding the first way of evaluation (max/min analysis), no single decomposition method consistently delivers the lowest overall memory footprint across all Conv layers, as the outcome depends on the specific layer shape. However, across the 14 CNNs studied in this work, Tucker achieves the lowest average overall memory footprint. This is primarily because Tucker results in fewer Conv layers, and therefore fewer FMs, compared to the other methods.

For example, in 2D Conv layers, Tucker produces three Conv layers, while the other methods yield four. This reduction in layer count leads to smaller FM memory consumption for Tucker. The advantage is even greater in 3D Conv layers: Tucker still uses only three layers, whereas alternative methods can produce up to five, further increasing Tucker’s memory efficiency.

When FMs occupy more memory than parameters (typically in cases with large spatial dimensions), Tucker shows a clear advantage in minimizing total memory usage. In FC layers, SVD and QR consistently yield the lowest overall memory footprint. Although T3F achieves lower parameter counts, its high FM memory makes it less efficient overall.

Regarding the second way of evaluation (constraint-based analysis), LRF methods yield significantly different overall memory values under fixed parameter or FLOPs constraints, depending on the layer's shape.

\noindent \textbf{FLOPs.}
Regarding the first way of evaluation (max/min analysis), all methods apart from T3F exhibit similar best and worst achievable FLOPs, with a maximum difference of only 1\% (Fig.~\ref{fig:spider_conv} and Fig.~\ref{fig:spider_fc}).

Regarding the constraint-based analysis, for a specific parameter count, CP yields the lowest FLOPs due to its use of depthwise convolution, followed by TT. In contrast, Tucker results in the highest FLOPs; however, the difference between Tucker and TT is minor.

T3F results in a higher FLOP count than SVD/QR due to its use of tensor contraction instead of conventional matrix-matrix multiplication.
Additionally, T3F introduces additional reshape layers, which increase memory accesses and further degrade the inference time~\cite{milad3}.

% \usepackage{multirow}
\begin{table}[t]
\caption{Evaluation in terms of ES for six Conv layers.}

\centering

\scalebox{0.84}{

\begin{tabular}{cccccccc}
\hline
\rowcolor[HTML]{DAE8FC}
\cellcolor[HTML]{DAE8FC} & \multicolumn{3}{c}{\cellcolor[HTML]{DAE8FC}All/Selected   LRF Solutions} & \cellcolor[HTML]{DAE8FC} & \multicolumn{3}{c}{\cellcolor[HTML]{DAE8FC}Solutions per step} \\ \cline{2-4} \cline{6-8}
\rowcolor[HTML]{DAE8FC}
\multirow{-2}{*}{\cellcolor[HTML]{DAE8FC}\begin{tabular}[c]{@{}c@{}}Layer\\ $(K_1,...,K_d,C,F)$\end{tabular}} & Tucker & CP & TT & \multirow{-2}{*}{\cellcolor[HTML]{DAE8FC}\begin{tabular}[c]{@{}c@{}}Comp.\\ ratio\end{tabular}} & Tucker & CP & TT \\ \hline
\cellcolor[HTML]{DAE8FC} &  &  &  & 85\% & 3 & 1 & 3 \\
\cellcolor[HTML]{DAE8FC} &  &  &  & 60\% & 2 & 1 & 2 \\
\multirow{-3}{*}{\cellcolor[HTML]{DAE8FC}1D-(3,512,1024)} & \multirow{-3}{*}{\begin{tabular}[c]{@{}c@{}}5.2E+5\\ 4.1E+5\end{tabular}} & \multirow{-3}{*}{\begin{tabular}[c]{@{}c@{}}1.5E+3\\ 1.0E+3\end{tabular}} & \multirow{-3}{*}{\begin{tabular}[c]{@{}c@{}}5.2E+5\\ 4.1E+5\end{tabular}} & 25\% & 2 & 1 & 2 \\ \hline
\cellcolor[HTML]{DAE8FC} &  &  &  & 85\% & 2 & 1 & 34 \\
\cellcolor[HTML]{DAE8FC} &  &  &  & 60\% & 2 & 1 & 142 \\
\multirow{-3}{*}{\cellcolor[HTML]{DAE8FC}2D-(3,3,256,512)} & \multirow{-3}{*}{\begin{tabular}[c]{@{}c@{}}1.3E+5\\ 1.2E+5\end{tabular}} & \multirow{-3}{*}{\begin{tabular}[c]{@{}c@{}}2.3E+3\\ 7.6E+2\end{tabular}} & \multirow{-3}{*}{\begin{tabular}[c]{@{}c@{}}1.0E+8\\ 5.5E+7\end{tabular}} & 25\% & 1 & 0 & 0 \\ \hline
\cellcolor[HTML]{DAE8FC} &  &  &  & 85\% & 3 & 1 & 483 \\
\cellcolor[HTML]{DAE8FC} &  &  &  & 60\% & 3 & 1 & 485 \\
\multirow{-3}{*}{\cellcolor[HTML]{DAE8FC}2D-(3,3,512,512)} & \multirow{-3}{*}{\begin{tabular}[c]{@{}c@{}}2.6E+5\\ 2.5E+5\end{tabular}} & \multirow{-3}{*}{\begin{tabular}[c]{@{}c@{}}4.6E+3\\ 2.2E+3\end{tabular}} & \multirow{-3}{*}{\begin{tabular}[c]{@{}c@{}}4.0E+8\\ 2.6E+8\end{tabular}} & 25\% & 3 & 1 & 720 \\ \hline
\cellcolor[HTML]{DAE8FC} &  &  &  & 85\% & 2 & 1 & 19 \\
\cellcolor[HTML]{DAE8FC} &  &  &  & 60\% & 3 & 1 & 11 \\
\multirow{-3}{*}{\cellcolor[HTML]{DAE8FC}2D-(5,5,96,256)} & \multirow{-3}{*}{\begin{tabular}[c]{@{}c@{}}2.4E+4\\ 2.4E+4\end{tabular}} & \multirow{-3}{*}{\begin{tabular}[c]{@{}c@{}}2.4E+3\\ 1.6E+3\end{tabular}} & \multirow{-3}{*}{\begin{tabular}[c]{@{}c@{}}1.1E+7\\ 8.8E+6\end{tabular}} & 25\% & 1 & 1 & 1 \\ \hline
\cellcolor[HTML]{DAE8FC} &  &  &  & 85\% & 2 & 1 & 57 \\
\cellcolor[HTML]{DAE8FC} &  &  &  & 60\% & 2 & 1 & 33 \\
\multirow{-3}{*}{\cellcolor[HTML]{DAE8FC}2D-(3,3,384,256)} & \multirow{-3}{*}{\begin{tabular}[c]{@{}c@{}}9.8E+4\\ 9.5E+4\end{tabular}} & \multirow{-3}{*}{\begin{tabular}[c]{@{}c@{}}2.3E+3\\ 1.3E+3\end{tabular}} & \multirow{-3}{*}{\begin{tabular}[c]{@{}c@{}}7.5E+7\\ 5.3E+7\end{tabular}} & 25\% & 2 & 1 & 22 \\ \hline
\cellcolor[HTML]{DAE8FC} &  &  &  & 85\% & 3 & 1 & 423 \\
\cellcolor[HTML]{DAE8FC} &  &  &  & 60\% & 3 & 1 & 115 \\
\multirow{-3}{*}{\cellcolor[HTML]{DAE8FC}3D-(3,3,3,32,32)} & \multirow{-3}{*}{\begin{tabular}[c]{@{}c@{}}1.0E+3\\ 1.0E+3\end{tabular}} & \multirow{-3}{*}{\begin{tabular}[c]{@{}c@{}}8.6E+2\\ 1.0E+2\end{tabular}} & \multirow{-3}{*}{\begin{tabular}[c]{@{}c@{}}9.4E+6\\ 3.1E+6\end{tabular}} & 25\% & 3 & 1 & 0 \\ \hline
\end{tabular}
}
\squeezeup{}
\squeezeup{}
\squeezeup{}
\squeezeup{}
\squeezeup{}
\squeezeup{}

\label{table:conv_solutions}
\end{table}

\noindent \textbf{Exploration space (ES).}
The ES represents the total number of solutions that can be generated by a given decomposition method.
For most methods (excluding T3F), the ES is determined by the number of unique rank configurations the method supports.
In contrast, T3F offers not only a variety of rank solutions but also a diverse range of tensor shape configurations (i.e., combination shapes), dramatically expanding its ES.

Table ~\ref{table:conv_solutions} and Table ~\ref{table:fc-solutions} present a comprehensive analysis of all valid LRF configurations for six Conv and three FC layers, respectively.
The selected layers reflect common configurations within the 14 selected CNNs (more than 80\% of the layers exhibit similar structural properties and decomposition responses).
Table~\ref{table:conv_solutions} and Table~\ref{table:fc-solutions} show the number of LRF configurations that result in lower parameters and FLOPs compared to the original layer. Solutions are grouped into three compression levels, i.e.,  low (25\%), medium (60\%), and high (85\%), to illustrate the trade-off space and diversity of viable options under practical constraints.

The higher the number of ranks, the greater the number of possible rank combinations, and thus, the larger the ES.
For Conv layers, TT yields the largest ES due to its use of the most rank values, followed by Tucker, which allows for multiple ranks (Table~\ref{table:conv_solutions}).
For FC layers, SVD and QR use a single rank, resulting in a small ES (Table~\ref{table:fc-solutions}).
On the other hand, T3F provides an exceptionally large ES, as it supports multiple ranks and a wide variety of tensor shapes and configurations~\cite{Novikov2015} (Table~\ref{table:fc-solutions}).
However, this versatility makes T3F a powerful tool for optimizing FC layers, enabling more diverse and efficient configurations.

The higher the ES, the longer the time needed to extract and process all possible solutions ('solution generation time' in Table~\ref{table:conv-time} and Table ~\ref{table:fc-time}).
The results indicate that, in some cases, specially in TT decomposition, the 'solution generation time' can take several hours. The 'solution generation time' is defined as the time needed to obtain all solutions and store them into memory. All runtime measurements are obtained on the system described in Section~\ref{sec:experimental-setup}.

% \usepackage{multirow}
\begin{table}[t]
\caption{Evaluation in terms of ES for three FC layers.}

\centering
% \scalebox{0.85}{

\begin{tabular}{cccccccc}
\hline
\rowcolor[HTML]{DAE8FC}
\cellcolor[HTML]{DAE8FC} & \multicolumn{3}{c}{\cellcolor[HTML]{DAE8FC}LRF solutions in overall} & \cellcolor[HTML]{DAE8FC} & \multicolumn{3}{c}{\cellcolor[HTML]{DAE8FC}Solutions per step} \\ \cline{2-4} \cline{6-8}
\rowcolor[HTML]{DAE8FC}
\multirow{-2}{*}{\cellcolor[HTML]{DAE8FC}\begin{tabular}[c]{@{}c@{}}Layer\\ shape\end{tabular}} & SVD & QR & T3F & \multirow{-2}{*}{\cellcolor[HTML]{DAE8FC}\begin{tabular}[c]{@{}c@{}}Comp.\\ ratio\end{tabular}} & SVD & QR & T3F \\ \hline
\cellcolor[HTML]{DAE8FC} &  &  &  & 85\% & 1 & 1 & 4 \\
\cellcolor[HTML]{DAE8FC} &  &  &  & 60\% & 1 & 1 & 1 \\
\multirow{-3}{*}{\cellcolor[HTML]{DAE8FC}(400,120)} & \multirow{-3}{*}{\begin{tabular}[c]{@{}c@{}}1E+2\\ 9E+1\end{tabular}} & \multirow{-3}{*}{\begin{tabular}[c]{@{}c@{}}1E+2\\ 9E+1\end{tabular}} & \multirow{-3}{*}{\begin{tabular}[c]{@{}c@{}}8E+5\\ 3E+4\end{tabular}} & 25\% & 1 & 1 & 0 \\ \hline
\cellcolor[HTML]{DAE8FC} &  &  &  & 85\% & 1 & 1 & 2 \\
\cellcolor[HTML]{DAE8FC} &  &  &  & 60\% & 1 & 1 & 3 \\
\multirow{-3}{*}{\cellcolor[HTML]{DAE8FC}(512,512)} & \multirow{-3}{*}{\begin{tabular}[c]{@{}c@{}}5E+2\\ 2E+2\end{tabular}} & \multirow{-3}{*}{\begin{tabular}[c]{@{}c@{}}5E+2\\ 2E+2\end{tabular}} & \multirow{-3}{*}{\begin{tabular}[c]{@{}c@{}}3E+6\\ 9E+4\end{tabular}} & 25\% & 1 & 1 & 0 \\ \hline
\cellcolor[HTML]{DAE8FC} &  &  &  & 85\% & 1 & 1 & 6 \\
\cellcolor[HTML]{DAE8FC} &  &  &  & 60\% & 1 & 1 & 2 \\
\multirow{-3}{*}{\cellcolor[HTML]{DAE8FC}(512,256)} & \multirow{-3}{*}{\begin{tabular}[c]{@{}c@{}}2E+2\\ 1E+2\end{tabular}} & \multirow{-3}{*}{\begin{tabular}[c]{@{}c@{}}2E+2\\ 1E+2\end{tabular}} & \multirow{-3}{*}{\begin{tabular}[c]{@{}c@{}}1E+6\\ 3E+4\end{tabular}} & 25\% & 1 & 1 & 0 \\ \hline
\end{tabular}
\squeezeup{}
\squeezeup{}
\squeezeup{}
\squeezeup{}
\squeezeup{}
\squeezeup{}

% }
\label{table:fc-solutions}
\end{table}

\noindent \textbf{Parameter and FLOPs Coverage Across the ES.}
It is important to note that the distribution of solutions in the ES is not uniform across the Memory-FLOPs space (Fig.~\ref{fig:spider_conv} and Fig.~\ref{fig:spider_fc}).
For example, T3F solutions only appear in high memory reduction scenarios, with none present when memory reduction is low (Fig.~\ref{fig:LRF_FBP_solutions}). In Fig.~\ref{fig:LRF_FBP_solutions}, SVD provides a higher memory coverage than T3F.
In Conv layers, all methods provide the same parameter coverage while in FC layers all methods provide the same FLOPs coverage.

\noindent \textbf{Flexibility.}
Flexibility, in this context, refers to the ability to adjust factorization independently across tensor dimensions, allowing for better adaptation to data structure and redundancy. For Conv layers, Tucker is the most flexible, supporting selective decomposition and separate rank values per dimension. TT also allows multiple ranks, making it more flexible than CP, which uses a single rank across all dimensions.
For FC layers, T3F offers greater flexibility than SVD and QR by supporting multiple ranks and various tensor reshaping options, unlike SVD and QR which rely on a single rank and fixed structure.

\noindent \textbf{Tensor/Matrix Decomposition process.}
The time required for tensor or matrix decomposition is a critical factor, often taking several hours in some decomposition methods (Table~\ref{table:conv-time}). Decomposition time varies significantly across methods due to differences in tensor size, selected rank(s), computational complexity, optimization strategies, and factorization processes.

% \usepackage{multirow}
\begin{table}[t]
\caption{Evaluation in terms of extraction and decomposition time for six Conv layers.}
\centering

\scalebox{0.82}{

\begin{tabular}{cccccccc}
\hline
\rowcolor[HTML]{DAE8FC}
\cellcolor[HTML]{DAE8FC} & \multicolumn{3}{c}{\cellcolor[HTML]{DAE8FC}Solution generation time (s)} & \cellcolor[HTML]{DAE8FC} & \multicolumn{3}{c}{\cellcolor[HTML]{DAE8FC}Decomposition time (s)} \\ \cline{2-4} \cline{6-8}
\rowcolor[HTML]{DAE8FC}
\multirow{-2}{*}{\cellcolor[HTML]{DAE8FC}\begin{tabular}[c]{@{}c@{}}Layer\\ $(K_1,...,K_d,C,F)$\end{tabular}} & Tucker & CP & TT & \multirow{-2}{*}{\cellcolor[HTML]{DAE8FC}\begin{tabular}[c]{@{}c@{}}Comp. \\ ratio\end{tabular}} & Tucker & CP & TT \\ \hline
\cellcolor[HTML]{DAE8FC} &  &  &  & 85\% & 51 & 3926 & 82 \\
\cellcolor[HTML]{DAE8FC} &  &  &  & 60\% & 229 & 20747 & 83 \\
\multirow{-3}{*}{\cellcolor[HTML]{DAE8FC}1D-(3,512,1024)} & \multirow{-3}{*}{39} & \multirow{-3}{*}{0.1} & \multirow{-3}{*}{48} & 25\% & 402 & - & 97 \\ \hline
\cellcolor[HTML]{DAE8FC} &  &  &  & 85\% & 66 & 192 & 29 \\
\cellcolor[HTML]{DAE8FC} &  &  &  & 60\% & 74 & 411 & 50 \\
\multirow{-3}{*}{\cellcolor[HTML]{DAE8FC}2D-(3,3,256,512)} & \multirow{-3}{*}{21.6} & \multirow{-3}{*}{0.4} & \multirow{-3}{*}{3724} & 25\% & 97 & - & - \\ \hline
\cellcolor[HTML]{DAE8FC} &  &  &  & 85\% & 146 & 687 & 40 \\
\cellcolor[HTML]{DAE8FC} &  &  &  & 60\% & 194 & 2495 & 45 \\
\multirow{-3}{*}{\cellcolor[HTML]{DAE8FC}2D-(3,3,512,512)} & \multirow{-3}{*}{31.7} & \multirow{-3}{*}{0.7} & \multirow{-3}{*}{28872} & 25\% & 181 & 3890 & 40 \\ \hline
\cellcolor[HTML]{DAE8FC} &  &  &  & 85\% & 51 & 889 & 28 \\
\cellcolor[HTML]{DAE8FC} &  &  &  & 60\% & 42 & 3049 & 8 \\
\multirow{-3}{*}{\cellcolor[HTML]{DAE8FC}2D-(5,5,96,256)} & \multirow{-3}{*}{4} & \multirow{-3}{*}{0.4} & \multirow{-3}{*}{598} & 25\% & 34 & 6554 & 22 \\ \hline
\cellcolor[HTML]{DAE8FC} &  &  &  & 85\% & 62 & 1462 & 9 \\
\cellcolor[HTML]{DAE8FC} &  &  &  & 60\% & 79 & 3572 & 20 \\
\multirow{-3}{*}{\cellcolor[HTML]{DAE8FC}2D-(3,3,384,256)} & \multirow{-3}{*}{11.1} & \multirow{-3}{*}{0.4} & \multirow{-3}{*}{3390} & 25\% & 103 & 6819 & 42 \\ \hline
\cellcolor[HTML]{DAE8FC} &  &  &  & 85\% & 1 & 25 & 1 \\
\cellcolor[HTML]{DAE8FC} &  &  &  & 60\% & 3 & 149 & 2 \\
\multirow{-3}{*}{\cellcolor[HTML]{DAE8FC}3D-(3,3,3,32,32)} & \multirow{-3}{*}{0.2} & \multirow{-3}{*}{0.2} & \multirow{-3}{*}{203} & 25\% & 2 & 277 & - \\ \hline
\end{tabular}
}

\squeezeup{}
\squeezeup{}
\squeezeup{}
\squeezeup{}
\squeezeup{}
\squeezeup{}

\label{table:conv-time}
\end{table}

Among tensor decomposition methods, TT is the fastest due to its non-iterative, SVD-based process. CP is the slowest, requiring time-consuming iterative optimization and often facing convergence issues, sometimes taking hours to decompose a single tensor. Tucker is also iterative but faster than CP, as we limit decomposition to two dimensions. All tensor decompositions were performed using the Tensorly library~\cite{tensorly}. In summary, TT is fastest, followed by Tucker, with CP being the slowest (Table~\ref{table:conv-time}); the difference between TT and Tucker is minor.

For FC layers, SVD and QR are more efficient than tensor methods, operating on simpler 2D matrices (Table~\ref{table:fc-time}). SVD is deterministic but computationally heavier for large layers due to quadratic complexity. QR is generally faster and non-iterative, though rank truncation may be needed for compression. T3F offers greater flexibility but incurs extra cost from tensorization, though still under one second.

\noindent \textbf{Key Observations: }

\begin{enumerate}
    \item Different LRF methods exhibit significant variations in terms of ES, parameter count, overall memory, FLOPs, parameter/FLOPs coverage and decomposition time.

    \item For a fixed parameter count, different LRF methods produce varying FLOPs depending on the layer shape (and vice versa). Similarly, the resulting FM memory (and thus overall memory) can differ significantly between methods.

    \item FC layers: SVD/QR is the best choice when low inference time or minimal memory footprint is required, as T3F incurs higher FLOPs and additional reshape layers ~\cite{milad3}. However, T3F offers higher ES and lower parameter counts, although it offers no solutions at low compression ratios. T3F may be preferable for ML engineers, as its multiple solutions per compression level enable greater potential for accuracy improvement.

    \item Conv Layers: All three methods yield similar results in terms of parameter count and FLOPs. However, CP is less suitable for ML engineers as it provides only one solution per compression ratio, and thus there is less potential for accuracy improvement. Additionally, CP is not ideal when decomposition time is an issue.

\end{enumerate}

% \usepackage{multirow}
\begin{table}[t]
\caption{Evaluation in terms of extraction and decomposition time for three FC layers.}
\centering

\scalebox{0.93}{

\begin{tabular}{cccccccc}
\hline
\rowcolor[HTML]{DAE8FC}
\cellcolor[HTML]{DAE8FC} & \multicolumn{3}{c}{\cellcolor[HTML]{DAE8FC}Solution generation time (s)} & \cellcolor[HTML]{DAE8FC} & \multicolumn{3}{c}{\cellcolor[HTML]{DAE8FC}Decomposition time (s)} \\ \cline{2-4} \cline{6-8}
\rowcolor[HTML]{DAE8FC}
\multirow{-2}{*}{\cellcolor[HTML]{DAE8FC}\begin{tabular}[c]{@{}c@{}}Layer\\ shape\end{tabular}} & SVD & QR & T3F & \multirow{-2}{*}{\cellcolor[HTML]{DAE8FC}\begin{tabular}[c]{@{}c@{}}Comp.\\ ratio\end{tabular}} & SVD & QR & T3F \\ \hline
\cellcolor[HTML]{DAE8FC} &  &  &  & 85\% & 0.01 & 0.01 & 0.09 \\
\cellcolor[HTML]{DAE8FC} &  &  &  & 60\% & 0.01 & 0.01 & 0.08 \\
\multirow{-3}{*}{\cellcolor[HTML]{DAE8FC}(400,120)} & \multirow{-3}{*}{0.1} & \multirow{-3}{*}{0.1} & \multirow{-3}{*}{57} & 25\% & 0.01 & 0.01 & - \\ \hline
\cellcolor[HTML]{DAE8FC} &  &  &  & 85\% & 0.11 & 0.03 & 0.09 \\
\cellcolor[HTML]{DAE8FC} &  &  &  & 60\% & 0.1 & 0.04 & 0.24 \\
\multirow{-3}{*}{\cellcolor[HTML]{DAE8FC}(512,512)} & \multirow{-3}{*}{0.1} & \multirow{-3}{*}{0.1} & \multirow{-3}{*}{313} & 25\% & 0.13 & 0.03 & - \\ \hline
\cellcolor[HTML]{DAE8FC} &  &  &  & 85\% & 0.05 & 0.02 & 0.1 \\
\cellcolor[HTML]{DAE8FC} &  &  &  & 60\% & 0.04 & 0.02 & 0.08 \\
\multirow{-3}{*}{\cellcolor[HTML]{DAE8FC}(512,256)} & \multirow{-3}{*}{0.1} & \multirow{-3}{*}{0.1} & \multirow{-3}{*}{98} & 25\% & 0.04 & 0.02 & - \\ \hline
\end{tabular}
}
\squeezeup{}
\squeezeup{}
\squeezeup{}
\squeezeup{}
\squeezeup{}
\squeezeup{}
\squeezeup{}

\label{table:fc-time}
\end{table}

\section{Proposed Methodology and Framework} \label{sec:methodology}

In this Section, the proposed methodology and framework is provided.
In Subsection~\ref{subsec:date-methodology}, we describe the proposed framework when applying a single decomposition method for the Conv layers and a single decomposition method for the FC layers.
In Subsection~\ref{subsec:combining-methods}, we introduce a post-processing step that applies hybrid decomposition, where different decomposition methods are employed in different CNN layers, further enhancing compression efficiency.

\subsection{Proposed Methodology and Framework} \label{subsec:date-methodology}

\begin{figure*}[htbp]
% \centerline{\includegraphics[width=1\textwidth]{figures/Methodology_similarity3 (1).png}}
\centerline{\includegraphics[width=1\textwidth]{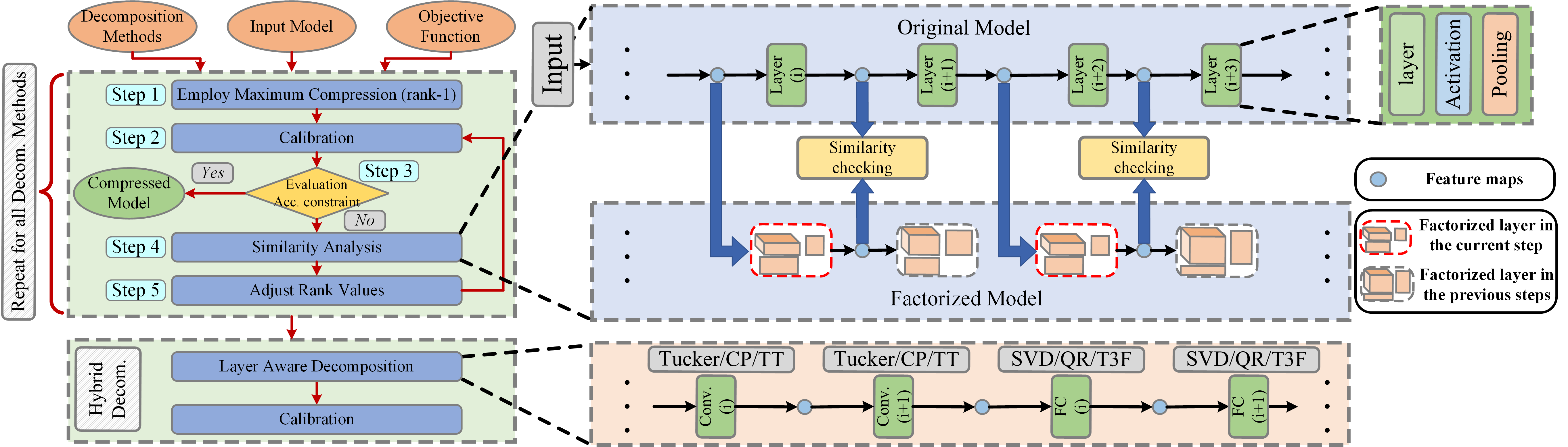}}
\caption{Proposed Methodology and Framework}

\squeezeup{}
\squeezeup{}
\squeezeup{}
\squeezeup{}
\squeezeup{}

\label{fig:methodology}
\end{figure*}

The main steps of the proposed methodology/framework are shown in the left part of Fig.~\ref{fig:methodology}. Each step is further explained below.
Initially, the user specifies the input model and the objective function, i.e., minimizing FLOPs, parameter count, or overall memory.
Next, the process shown in the top-left box of Fig.~\ref{fig:methodology} is repeated using a fixed LRF method for the Conv layers and a different fixed method for the FC layers.

We start by selecting a subset of layers as the target for factorization. Although the proposed methodology can be applied to all layers, focusing on a subset allows for more efficient convergence to an efficient solution. In this study, we choose to factorize 90\% of the layers (exclude the small layers with respect to the objective function). This approach balances reduced computational cost (e.g., number of operations) and memory footprint with preserved model accuracy.

\noindent\textbf{Step1. Employ Maximum Compression (rank-1):}
Since the aim of the proposed methodology is to minimize the objective function, we begin by selecting the optimal baseline: a model in which all layers have a rank of one.
This initial configuration consistently yields the lowest objective value, as it results in the smallest number of parameters and FLOPs among all factorized model variants.

\noindent\textbf{Step2. Calibration:}
To compensate for the accuracy loss caused by factorization, each solution requires a calibration phase, which is the most time-consuming part of the LRF process. To address this problem, we avoid re-training or iteratively fine-tuning the model after each layer is factorized. Instead, we employ a one-shot fine-tuning approach, where the factorized model is fine-tuned for a limited number of epochs, e.g., 10 epochs, only once, after all target layers have been factorized. As a result, the calibration time is significantly reduced.
In the results Section~\ref{sec:results} we show that the proposed one-shot fine-tuning approach is on average 8× faster compared to iterative fine-tuning.

\noindent\textbf{Step3. Evaluation:}
After fine-tuning the factorized model, its performance is evaluated. This can be assessed in various ways, e.g., validation loss, validation accuracy, or a combination of metrics. In this study, we use validation accuracy to compare the factorized model against the original model. Additionally, we apply a threshold to determine the acceptable performance difference between the factorized and original models. Specifically, we set a user-defined threshold (in this paper is set to 1.5\%), allowing for up to a 1.5\% drop in accuracy. If the factorized model meets this accuracy constraint, the process stops and returns the factorized model. If the factorized model fails to meet the accuracy requirement, the process proceeds to the next steps to explore alternative solutions.

\noindent\textbf{Step4. Similarity analysis:}
After fine-tuning, if the factorized model does not meet the input accuracy constraint, we must reduce the compression ratio by increasing the rank values. Since each layer impacts accuracy differently, we face a question; how to adjust each layer's ranks to maximize overall compression while maintaining accuracy close to the original model? Previous methods~\cite{VBMF, bayesopt} consider layers' weights individually without accounting for their interactions within the model. Another method~\cite{LRF13} focuses on the similarity between the factorized and original weights; this approach is time-consuming, because it requires reconstructing the weights from the factorized components and also overlooks layer interactions and calibration effects. To address these issues, we propose a novel similarity-based strategy that compares feature maps, rather than weights, using cosine similarity as the metric.

The proposed similarity strategy focuses on feature maps rather than weight tensors or matrices, providing a better understanding of each layer's impact on model accuracy. Furthermore, we consider feature maps after subsequent operations such as activation, pooling, and normalization.

To obtain the feature maps, we randomly select a subset of training data (1,000 samples in this study), feed them into the original model, and save the resulting feature maps. To calculate the similarity between the new factorized layers and the original ones, we follow this procedure (illustrated with blue arrows in the right part of Fig.~\ref{fig:methodology}): each factorized layer receives the corresponding feature map from the original model as input, and its output is compared against the feature map of the original layer to determine similarity.

This similarity measure is used to assess the impact of each factorized layer on the model's overall accuracy. The rationale for this is twofold: first, to ensure that similarity is not influenced by previously factorized layers, and second, to identify which layers have the highest effect on model's accuracy.
For example, consider two factorized layers: the first has a high impact on accuracy, while the second has a low impact. In this scenario, our adaptive method will leverage this information, applying a higher compression rate to the second layer. Finally, the cosine similarity between two vectors is calculated as:
\begin{equation}
\textrm{cosine similarity}=\dfrac{\textbf{A.B}}{\lVert \textbf{A} \rVert \lVert\textbf{B} \rVert}
 \label{eq:cosine}
\end{equation}
\noindent where $\textbf{A.B}$ is the dot product of the vectors, and $\lVert \textbf{A} \rVert$ and $\lVert\textbf{B} \rVert$ are the Euclidean norms of \textbf{A} and \textbf{B}, respectively.

\begin{figure*}[tbp]
% \centerline{\includegraphics[trim={0 0 0 1mm}, clip,width=0.99\textwidth]{figures/Conv_Parameter_Reduction_Conv_Methods_Step2_Solution3.png}}
\centerline{\includegraphics[trim={1mm 0 0 0}, clip,width=0.99\textwidth]{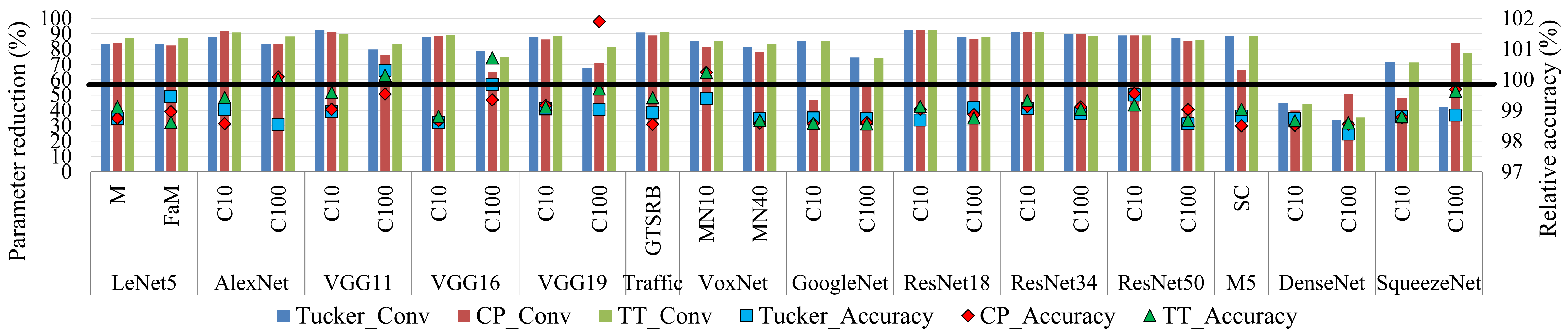}}

\squeezeup{}
\squeezeup{}
\squeezeup{}

\caption{Comparison with the non-factorized model for Conv layers, using \textit{step-size=2} and \textit{max-sol=10}.}

\squeezeup{}
\squeezeup{}
\squeezeup{}
\squeezeup{}
\squeezeup{}
\squeezeup{}

\label{fig:compared_to_original_parameter_conv}
\end{figure*}

\begin{figure}[tbp]
% \centerline{\includegraphics[trim={0 0 0 1mm}, clip,width=0.99\textwidth]{figures/FC_Parameter_Reduction_FC_Methods_Step2_Solution10.png}}
\centerline{\includegraphics[trim={1mm 0 0 1mm}, clip,width=0.49\textwidth]{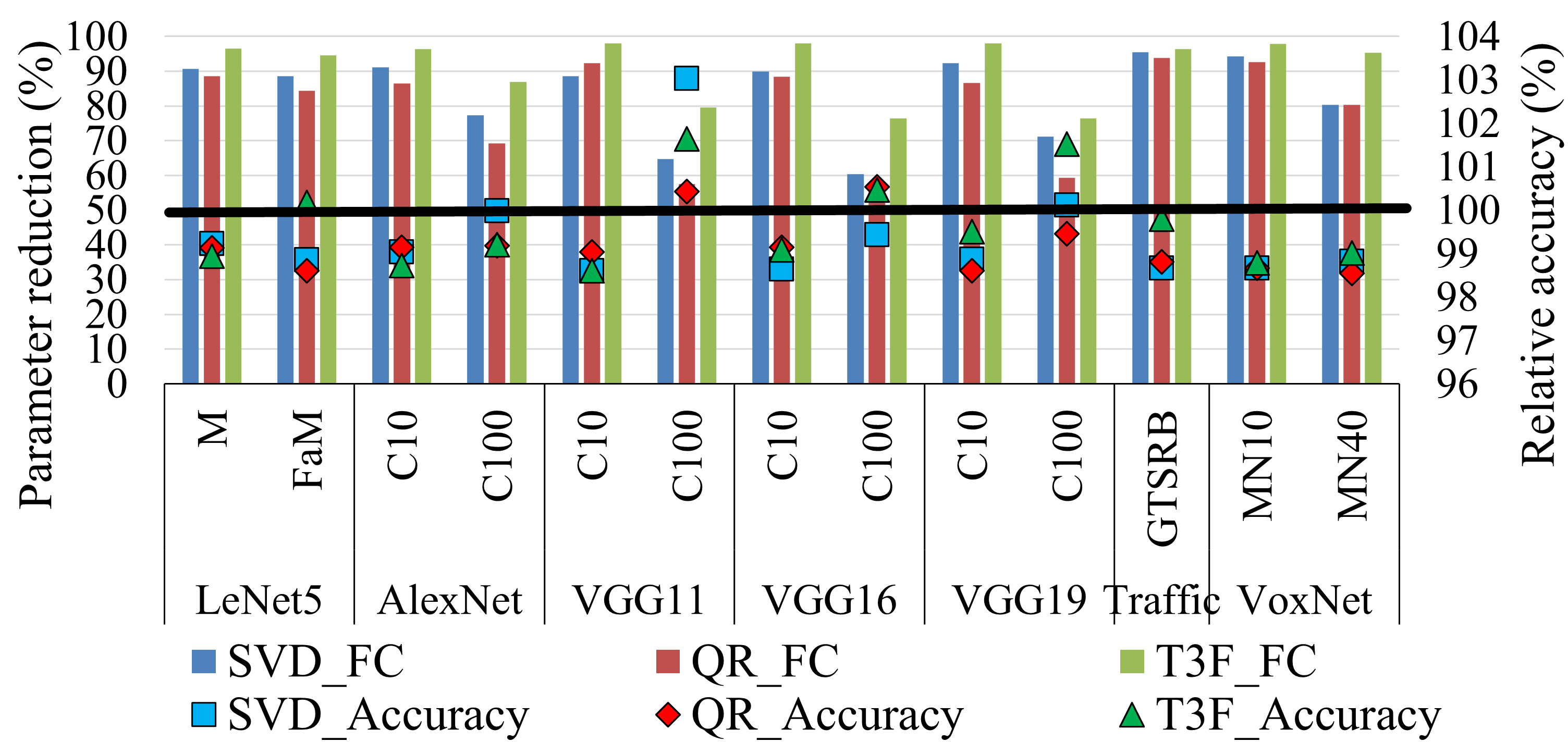}}

\squeezeup{}
\squeezeup{}
\squeezeup{}

\caption{Comparison with the non-factorized model for FC layers, using \textit{step-size=2} and \textit{max-sol=10}.}

\squeezeup{}
\squeezeup{}
\squeezeup{}

\label{fig:compared_to_original_parameter_fc}
\end{figure}

\noindent\textbf{Step5. Adjust Rank Values:}
After calculating the cosine similarity between the factorized layers and their original counterparts, we must determine whether to retain the factorized layer as it is or adjust their ranks based on a threshold.

Specifically, if the average similarity exceeds 0.92\footnote{The threshold is set to 0.96 for the layers in the non-sequential parts of the model}, we keep the current rank and solution for this layer (i.e., this is the final rank for this layer).
If the average similarity is below 0.92, we increase the rank based on a predefined compression step (\textit{step-size}).
In the previous work~\cite{milad-date}, a fixed step-size of 5\% is utilized, where the compression ratio of the target layers was reduced by 5\% in each iteration. However, in this study, in order to analyze the impact of this parameter on both the factorization time and the resulting compression ratio, three different step-size values: 2\%, 5\%, and 10\% are employed. The results corresponding to these step-size values are presented and discussed in Section VII.

The rationale for using a predefined compression step to increase the rank is as follows.
Given that the design space is vast (even for a single layer) and grows exponentially when considering multiple layers, evaluating all LRF solutions becomes impractical, as each solution must undergo a calibration phase.
Therefore, we introduce the \textit{step-size} approach to select a subset of LRF solutions for each layer and prune the design space.
The chosen \textit{step-size} value introduces a trade-off between achieving a higher compression ratio and processing time; a smaller step size results in more LRF candidates for a layer, necessitating more time for evaluation.

It is important to note that in certain decomposition methods, such as TT, multiple LRF solutions may exist for a given compression ratio.
In~\cite{milad-date}, three solutions were used for each compression ratio: i) the solution with the minimum FLOPs, ii) the solution with the maximum FLOPs, and iii) the solution with the same rank values across all dimensions. The first two solutions were selected based on the reasoning outlined in~\cite{milad2}, while the third was chosen for its ability to achieve lower approximation error, as discussed in~\cite{NAS}.

In this work, we extend the analysis by evaluating the effect of using different number of solutions (let \textit{max-sol}) on the compression ratio and model accuracy. Specifically, we consider three different scenarios: i) a single solution with the minimum FLOPs, ii) three solutions as described above, and iii) ten solutions. The latter scenario includes the three predefined solutions and seven additional randomly selected ones. The results for these scenarios are presented and analyzed in the results section.

After selecting the next solution, we return to step 2 to re-calibrate the factorized model. This process is repeated until a configuration that meets the accuracy constraints is found. This methodology ensures that layers with a greater impact on the overall accuracy (more sensitive layers) are compressed less, while layers with a lesser impact (less sensitive layers) are compressed more.

\subsection{Hybrid Decomposition - Combining Multiple Methods} \label{subsec:combining-methods}

In this subsection, the proposed framework is enhanced with a post-processing step that combines the six LRF methods into a hybrid decomposition approach. This is feasible due to the generality and flexibility of the proposed framework.

Combining different LRF methods on a layer-by-layer basis within a single model offers several benefits, allowing each layer to be compressed using the most suitable decomposition strategy based on its structure and computational characteristics. Consequently, it leads to lower overall memory usage and fewer FLOPs.

For example, Tucker decomposition works well for layers with large feature maps, as it can capture the interactions between different dimensions effectively (compresses along multiple dimensions of the tensor). Similarly, CP decomposition can provide advantages in certain layers, although it comes with longer decomposition times due to convergence challenges. By analyzing these layer-specific characteristics, different LRF methods can be strategically combined within the same model to maximize compression while maintaining accuracy.

The hybrid decomposition post-processing step is explained hereafter.
First, the proposed framework is applied separately using all three Conv decomposition methods and all three FC methods, resulting in nine combinations.
Next, the best-performing method is identified and employed for each layer based on the target objective function.
The resulting hybrid model is fine-tuned to recover any lost accuracy. If the calibration process does not achieve the required accuracy, we apply additional fine-tuning.
The user can also exclude specific methods from the framework. For example, CP decomposition can be skipped due to its high decomposition time.

Finally, it is important to highlight that since LRF methods preserve the input and output feature map shapes, they maintain compatibility with adjacent layers, ensuring seamless integration across the network.
Moreover, as the performance of each LRF method has already been evaluated for individual layers, the post-processing step involves systematically analyzing layer-wise results to select the most effective method for each layer. This structured selection process provides a practical pathway for optimally integrating different LRF methods, maximizing compression while maintaining accuracy and most importantly without introducing additional computational complexity.

\begin{figure*}[tbp]
% \centerline{\includegraphics[trim={0 0 0 1mm}, clip,width=0.99\textwidth]{figures/parameter_flops_reduction_avg_all_models_Convpng}}
\centerline{\includegraphics[trim={0 0 0 1mm}, clip,width=1\textwidth]{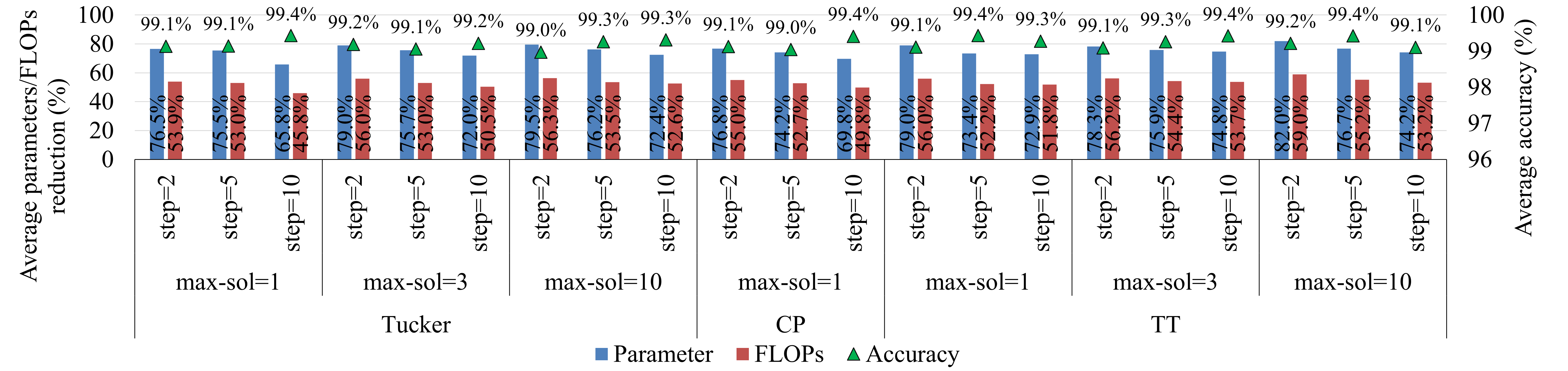}}

\squeezeup{}
\squeezeup{}
\squeezeup{}

\caption{Evaluating the impact of \textit{step-size} and \textit{max-sol} values on the proposed methodology for Conv layers }

\squeezeup{}
\squeezeup{}
\squeezeup{}

\label{fig:avg_conv_reduction}
\end{figure*}

\begin{figure}[tbp]
% \centerline{\includegraphics[trim={0 0 0 1mm}, clip,width=0.99\textwidth]{figures/parameter_flops_reduction_avg_all_models_fc.png}}
\centerline{\includegraphics[trim={1mm 0 0 0}, clip,width=0.49\textwidth]{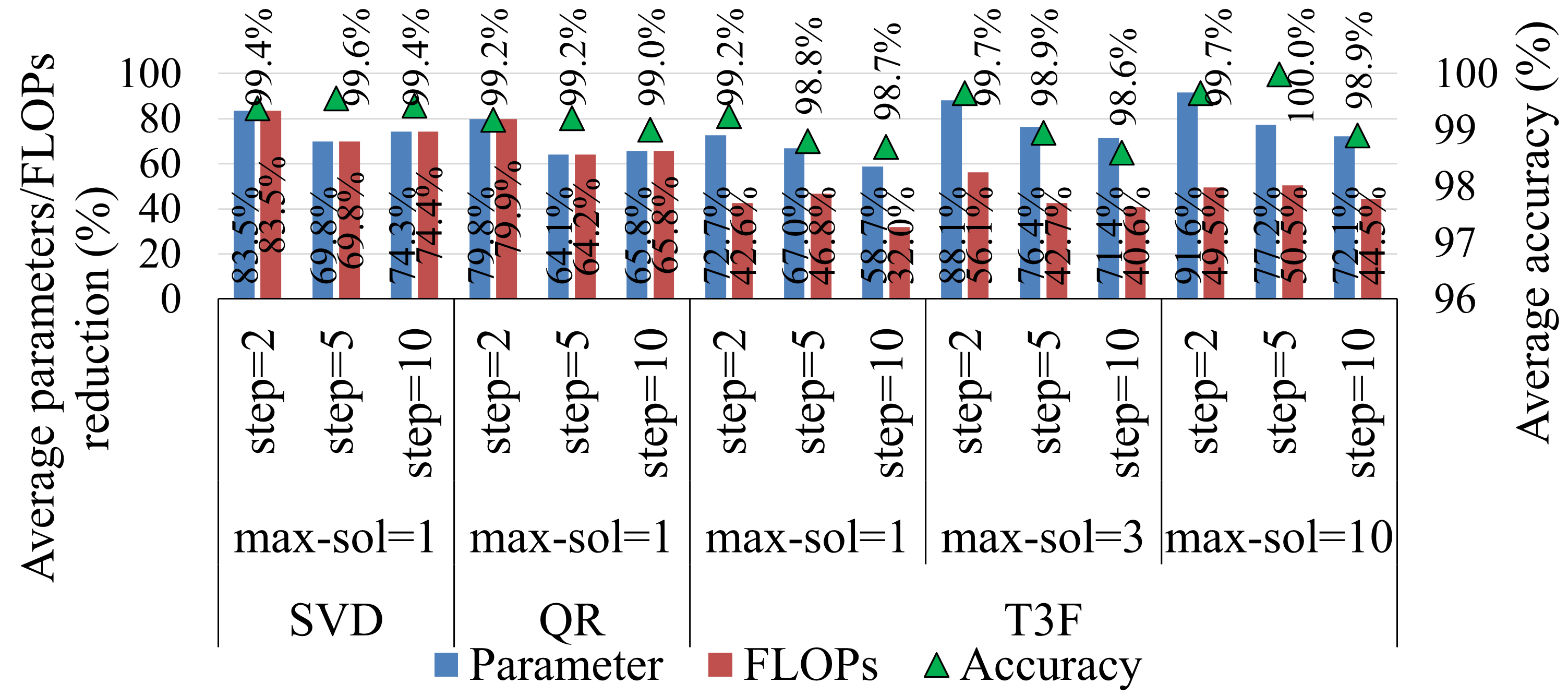}}

\squeezeup{}
\squeezeup{}
\squeezeup{}

\caption{Evaluating the impact of \textit{step-size} and \textit{max-sol} values on the proposed methodology for FC layers}

\squeezeup{}
\squeezeup{}
\squeezeup{}

\label{fig:avg_fc_reduction}
\end{figure}

%%%%%%%%%%%%%%%%%%%%%%%%%%%%%%%%%%%%%%%%%%%%%%%%%%%%%%%%%%%%%%%%%%%%%%%%%%%%%%%%%%%%%%%%%%%%%%%%%%%%%%%%%%%%%%%%%%%%%%%%%%%%%%%%%%%%%%%%%%%%%%%%%%%%%%%%%%%%%
\section{Experimental Setup}\label{sec:experimental-setup}

To demonstrate the effectiveness of our approach, we evaluated it using 14 CNN models with different basic blocks i.e., i) sequential-blocks: LeNet5, AlexNet, VGG11/16/19, Traffic Sign, VoxNet (3D Conv), M5 (1D Conv) ii) residual-blocks: ResNet-18/34/50, iii)inception-blocks: GoogleNet, iv) fire-blocks: SqueezeNet, and v) dense-blocks: DenseNet, on different datasets i.e., MNIST (M), Fashion-MNIST (FaM), ModelNet10 (MN10), ModelNet40 (MN40), Speech Commands (SC), GTSRB, CIFAR10 (C10), and CIFAR100 (C100).
All models are trained from scratch for 100 epochs using random weights, Adam optimizer by setting the initial learning rate to 1e-4. In addition, the ReduceLROnPlateau learning rate scheduler is used and configured with a patience parameter of 10 epochs, a factor of 0.1, and a batch size equals to 32. The factorized models are fine-tuned for 10 epochs using the entire training dataset. The validation accuracy is used to assess the performance of factorized models.
In all cases, we set a 1.5\% accuracy drop constraint as an acceptable accuracy degradation. Also, in order to streamline the process, we used the same similarity thresholds for all models i.e., 0.92 for sequential blocks and 0.96 for non-sequential blocks.

As there are no existing LRF tools for direct comparison, we evaluate the proposed methodology against the following approaches:

\begin{enumerate}
    \item {\bf Non-Factorized model}. % In all cases, we report parameter reduction, FLOPs reduction, and accuracy loss of the factorized models wrt. the corresponding original model.

    \item {\bf Variational Bayesian Matrix Factorization (VBMF)~\cite{VBMF}:} VBMF is a probabilistic approach for matrix factorization, which leverages Variational Bayesian inference to estimate the distributions of the latent factors and noise in the data. The advantage of VBMF over other matrix factorization techniques is its ability to automatically determine the rank. Given that our approach can be viewed as a rank selection method, comparing it to VBMF is a fair and relevant evaluation. In contrast to our rank selection methodology, which is based on feature map similarity rather than weights, VBMF automatically estimates the rank of each layer based on its weights.

    \item {\bf Filter-Based Pruning (FBP)~\cite{PR4}:} FBP is a well-known pruning technique used to reduce the parameters of CNNs by removing entire filters (or channels) from the Conv layers. Filters are pruned based on certain criteria to determine the ones contributing the least to the network's performance. In this work, we guide the FBP process using three different metrics (L1-norm, L2-norm, and Geometric Median Distance (GMD)~\cite{PR4}). For a fair comparison, the metric that offers the highest compression ratio, while keeping the accuracy drop to less than 1.5\%, is selected.

\end{enumerate}

All experiments were conducted using Tensorflow 2.15, Python 3.9.18, Tensorly 0.8.1 (for tensor decomposition), and Numpy 1.24.0 (for matrix decomposition). The experiments were carried out on a system with Ubuntu 22.04 OS and equipped with an Intel Xeon Silver 4309Y CPU at 2.80 GHz. It includes an NVIDIA A40 GPU and 256 GB RAM.

%%%%%%%%%%%%%%%%%%%%%%%%%%%%%%%%%%%%%%%%%%%%%%%%%%%%%%%%%%%%%%%%%%%%%%%%%%%%%%%%%%%%%%%%%%%%%%%%%%%%%%%%%%%%%%%%%%%%%%%%%%%%%%%%%%%%%%%%%%%%%%%%%%%%%%%%%%%%%

\section{Experimental Results}\label{sec:results}

\noindent \textbf{Comparison with the non-factorized model.}
Fig.~\ref{fig:compared_to_original_parameter_conv} and Fig.~\ref{fig:compared_to_original_parameter_fc} present the evaluation results against the original, non-factorized model, for the Conv and FC layers, respectively. The first graph is normalized to the Conv part, the second to the FC part.

As shown in Fig.~\ref{fig:compared_to_original_parameter_conv}, the proposed methodology achieves an average parameter reduction of 79.5\% in the Conv part (up to 92.3\% for VGG11 and ResNet18 on the C10 dataset), 76.8\% (up to 92.3\% for ResNet18 on the C10 dataset), and 82\% (up to 92.3\% for ResNet18 on the C10 dataset) for Tucker, CP, and TT decompositions, respectively.
As is evident from Fig.~\ref{fig:compared_to_original_parameter_conv}, in most cases, the three decomposition methods achieve similar compression ratios, with differences of less than 5\%. However, the method yielding the highest compression ratio varies across models. In certain cases, significant differences in compression ratios are observed. Specifically, in GoogleNet on C10 and C100, M5 on the SC dataset, and SqueezeNet on C10, Tucker and TT decompositions outperform CP decomposition. Conversely, in DenseNet on C100, CP decomposition surpasses Tucker and TT decompositions. Additionally, in SqueezeNet on C100, TT and CP decompositions achieve higher compression ratios than Tucker decomposition.

Furthermore, in certain cases, the factorized models exhibit improved accuracy compared to their original counterparts. Specifically, this accuracy enhancement is observed in VGG11 on C100 dataset with Tucker decomposition; AlexNet and VGG19 on C100 dataset, as well as VoxNet on MN10 dataset, with CP decomposition; and VGG11 and VGG16 on C100 dataset, along with VoxNet on MN10 dataset, with TT decomposition.

For the FC layers in Fig.~\ref{fig:compared_to_original_parameter_fc}, we show the parameter reduction achieved using SVD, QR, and T3F decomposition methods, focusing only on the FC part of the model. Models without a substantial FC component, such as those with only a final classification layer, are excluded, as matrix decomposition is not meaningful in those cases.
As Fig.~\ref{fig:compared_to_original_parameter_fc} indicates, the proposed methodology achieves an average parameter reduction in FC part of 83.5\% (up to 95.5\% in Traffic on GTSRB dataset), 79.8\% (up to 93.9\% in Traffic on GTSRB dataset), and 91.6\% (up to 98\% in VGG11, VGG16, and VGG16 on C10 dataset) for SVD, QR, and T3F decompositions, respectively.

As illustrated in Fig.~\ref{fig:compared_to_original_parameter_fc}, unlike Conv methods where no single approach consistently outperforms the others, T3F decomposition consistently surpasses SVD and QR in FC layers across all cases, achieving up to 20\% higher compression in certain instances. Additionally, with the exception of VGG11 on the C10 dataset, where QR decomposition yields slightly better results, SVD generally outperforms QR across all evaluated cases.
Furthermore, like Conv methods in certain cases, the factorized models exhibit improved accuracy compared to their original counterparts.

\begin{figure}[t]

\squeezeup{}
\squeezeup{}
\squeezeup{}
\squeezeup{}
\squeezeup{}
\squeezeup{}

\centerline{\includegraphics[trim={1mm 0 0 1mm}, clip,width=0.49\textwidth]{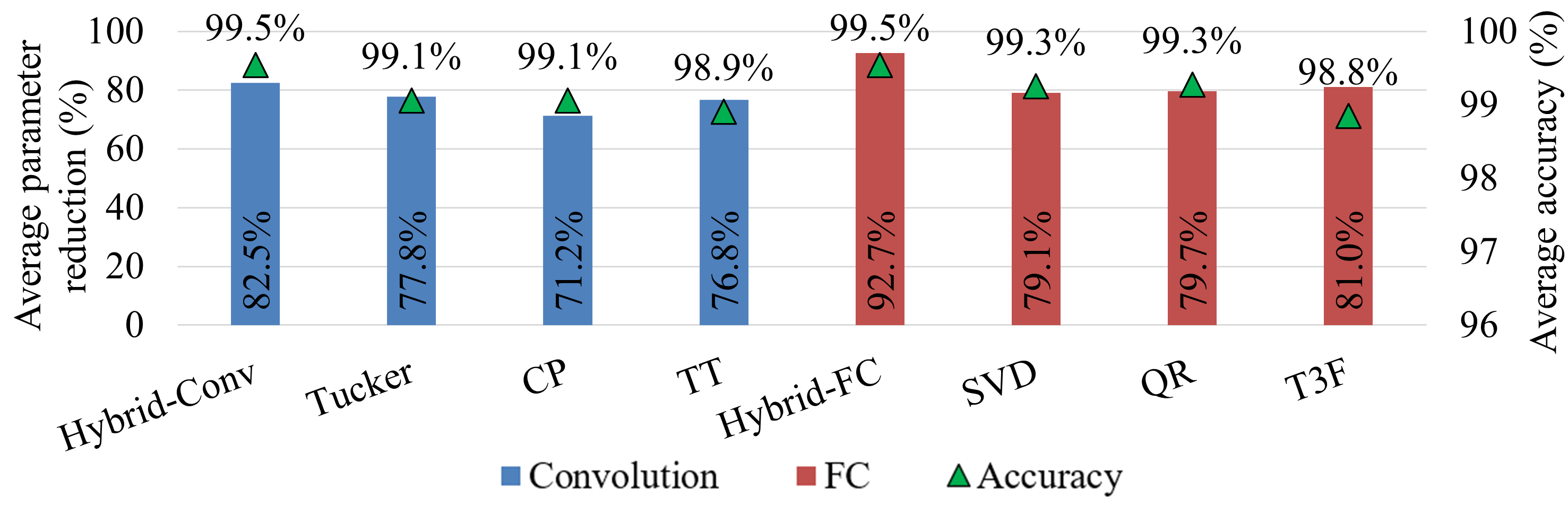}}

\squeezeup{}
\squeezeup{}
\squeezeup{}

\caption{Evaluation of hybrid decomposition when the objective function is parameter count}

\squeezeup{}
\squeezeup{}
\squeezeup{}

\label{fig:combining_parameter}
\end{figure}

\noindent \textbf{Effect of Threshold Selection on Methodology Performance.}
As discussed in Section~\ref{subsec:date-methodology}, our methodology relies on two key parameters: \textit{step-size} and \textit{max-sol}. The \textit{step-size} controls how the rank is increased when a layer fails to meet the similarity threshold, while \textit{max-sol} sets the maximum number of solutions explored per step, since some decomposition methods yield multiple candidates. In this subsection, we assess the effect of these parameters, using \textit{step-size} values of 2, 5, and 10, and \textit{max-sol} values of 1, 3, and 10.

Fig.~\ref{fig:avg_conv_reduction} presents the average parameter and FLOPs reductions (bars) across all studied models and datasets, along with the average relative accuracy (triangle marks), for all nine scenarios in the Conv layers using Tucker, CP, and TT decompositions.
Note that for CP, there is only one configuration related to \textit{max-sol}, as each LRF solution is determined by a single rank value.
As observed in Fig.~\ref{fig:avg_conv_reduction}, for a fixed number of solutions per step, there is an inverse relationship between parameter/FLOPs reduction and step-size. Specifically, when the step-size is set to two, the highest compression ratio is achieved. This occurs because a smaller step-size enables a smoother rank increase, providing more intermediate solutions to explore, which increases the likelihood of achieving higher compression.

Similarly, for a fixed step-size, the \textit{max-sol} generally exhibits the same trend: higher compression is expected when selecting more solutions per step. For instance, when the step-size is two, selecting three solutions typically results in higher compression than selecting one solution. Likewise, when the step-size is 10, selecting 10 solutions tends to yield higher compression than selecting three solutions. However, two exceptions are observed in TT decomposition.

Fig.~\ref{fig:avg_fc_reduction} presents a similar analysis, for FC layers.
Similar to CP decomposition, in both SVD and QR always \textit{max-sol=1}, as there is a single rank value.
As shown in Fig.~\ref{fig:avg_conv_reduction}, there is an inverse relationship between \textit{step-size} / \textit{max-sol} and the achieved compression ratio.
However, when considering FLOPs reduction instead of parameter reduction in T3F decomposition, this trend does not hold (Fig.~\ref{fig:avg_fc_reduction}). This deviation arises from the fact that, unlike SVD and QR, there is no linear relationship between parameter count and FLOPs in T3F ~\cite{milad3}.

\begin{figure}[t]

\squeezeup{}
\squeezeup{}
\squeezeup{}
\squeezeup{}
\squeezeup{}
\squeezeup{}

\centerline{\includegraphics[trim={0 0 0 1mm}, clip,width=0.49\textwidth]{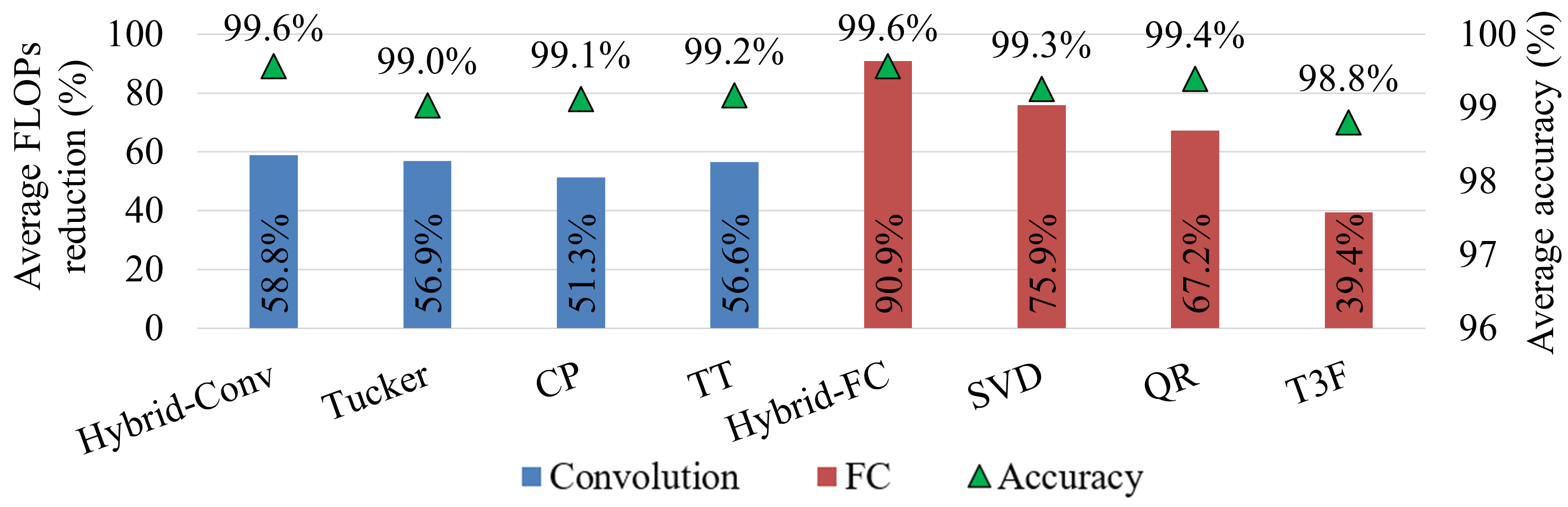}}

\squeezeup{}
\squeezeup{}
\squeezeup{}

\caption{Evaluation of hybrid decomposition when the objective function is FLOPs}

\squeezeup{}
\squeezeup{}
\squeezeup{}

\label{fig:combining_flops}
\end{figure}

\noindent \textbf{Hybrid Decomposition.}
In this Subsection we evaluate the post-processing step that combines the six LRF methods.
For clarity and focus, this subsection evaluates the hybrid method against Conv and FC methods separately, when the objective function is parameter reduction (Fig.\ref{fig:combining_parameter}) and FLOPs reduction (Fig.\ref{fig:combining_flops}).
The results are compared with the best performing configurations of other decomposition methods.
As expected, the Hybrid approach outperforms the individual methods, achieving a parameter reduction improvement ranging from 4.7\% to 11.3\% in the Conv layers and from 11.7\% to 13.6\% in the FC layers.
Furthermore, the hybrid model exhibits superior accuracy compared to other methods, demonstrating the effectiveness of this approach.

A similar analysis is presented in Fig.~\ref{fig:combining_flops}, when the objective is to maximize FLOPs reduction.
As expected, the Hybrid model outperforms the other decomposition methods, achieving from 1.9\% to 7.5\% in Conv layer and from 15\% to 51.5\% in FC layers FLOPs reduction.
Likewise for this objective, the hybrid model exhibits superior accuracy compared to other methods, demonstrating the effectiveness of this approach.

%%%%%%%%%%%%%%%%%%%%%%%%%%%%%%%%%%%%%%%%%%%%%%%%%%%%%%%%%%%%%%%
\begin{figure*}[tbp]
\centerline{\includegraphics[trim={1mm 0 0 1mm}, clip,width=1\textwidth]{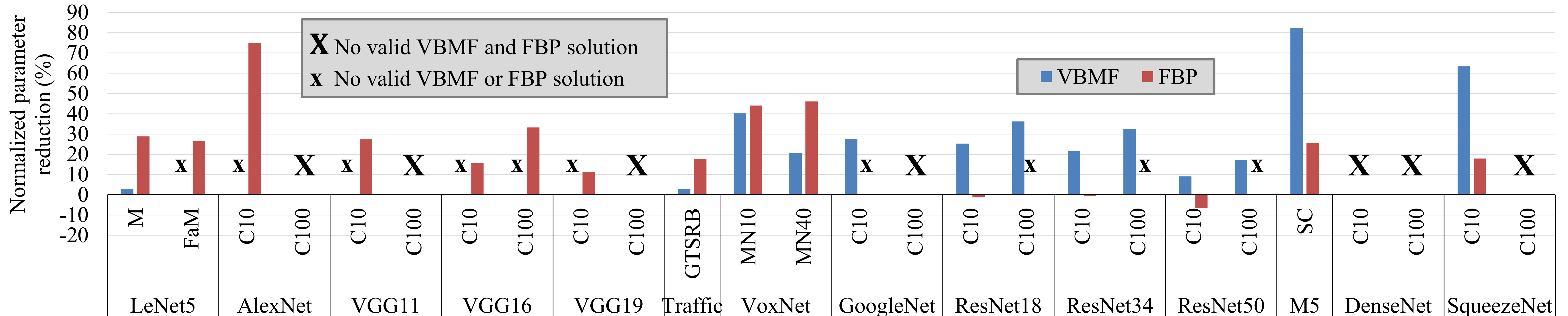}}

\squeezeup{}
\squeezeup{}
\squeezeup{}

\caption{Evaluation with VBMF and FBP. Tucker and SVD were used in all cases.}

\squeezeup{}
\squeezeup{}
\squeezeup{}

\label{fig:compared_to_VBMF_FBP}
\end{figure*}

\noindent \textbf{Comparison against VBMF and FBP.}
This subsection evaluates the proposed framework in Subsection~\ref{subsec:date-methodology} with VBMF and FBP.
We used Tucker decomposition for Conv layers with step-size=5 and max-sol=3, and SVD for FC layers with a step-size=2.
In all cases, the target layers, fine-tuning epochs, and accuracy drop constraints remain consistent across all studied methods. For FBP, three different filter pruning metrics are considered and the best-performing one is selected.
For VBMF, we use Tucker decomposition for Conv layers and SVD for FC layers.
Figure~\ref{fig:compared_to_VBMF_FBP} presents the relative parameter reduction achieved by our method over VBMF and FBP. Due to space constraints, only parameter reduction is shown, as FLOPs reduction exhibit similar trends. A capital “X” indicates the cases where both VBMF and FBP fail to meet accuracy constraints, while a lowercase “x” marks cases where at least one fails.

As seen in Fig.~\ref{fig:compared_to_VBMF_FBP}, VBMF and FBP fail to meet accuracy constraint in 13 and 11 out of 26 cases, respectively, demonstrating the robustness of our method. In all remaining cases, our approach consistently achieves higher compression ratios than VBMF, with up to 60\% additional compression (e.g., SqueezeNet on CIFAR-10). This advantage stems from our methodology’s ability to account for inter-layer dependencies, unlike VBMF, which compresses each layer independently.

A similar trend is observed when comparing our approach against FBP. The fundamental distinction is that LRF approximates layer weights, whereas FBP  remove network components (e.g., filters). Our approach outperforms FBP by 12\% to 75\% in most cases. However, for ResNet18/34/50 on CIFAR-10, FBP achieves up to 8\% higher compression; this is because FBP inherently benefits from implicitly compressing the next layer. Notably, in these cases, our method still reduces parameters by over 85\% relative to the original model (Fig.~\ref{fig:compared_to_original_parameter_conv}).

%%%%%%%%%%%%%%%%%%%%%%%%%%%%%%%%%%%%%%%%%%%%%%%%%%%%%%%%%%%%%%%
\begin{figure}[tbp]
\centerline{\includegraphics[trim={0 0 0 0mm}, clip,width=0.49\textwidth]{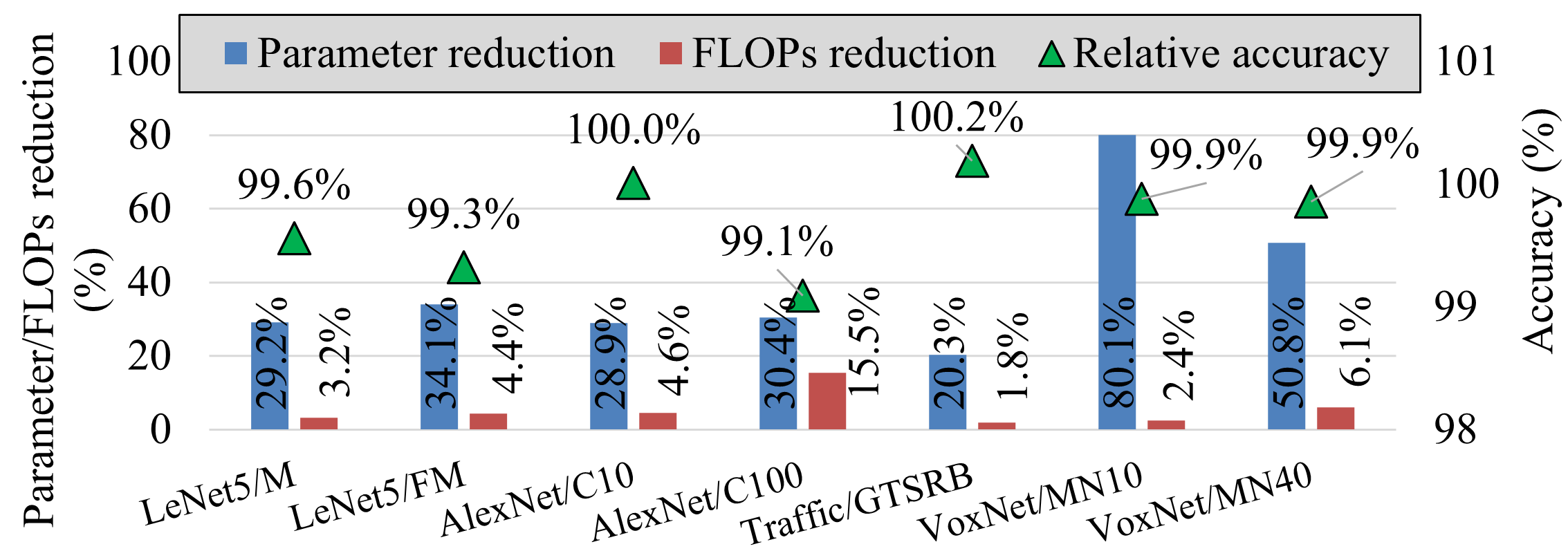}}

\squeezeup{}
\squeezeup{}
\squeezeup{}

\caption{Comparison of LRF vs. LRF+FBP. FBP is employed on the last Conv layer only.}

\squeezeup{}
\squeezeup{}
\squeezeup{}

\label{fig:lrf_fbp}
\end{figure}

\noindent \textbf{Compatibility with Other Compression Techniques: Applying FBP on Top of LRF.}
A key strength of our approach is its compatibility with additional compression methods, such as FBP. As shown in Fig.~\ref{fig:lrf_fbp}, applying FBP on top of the proposed LRF method yields further compression benefits. In this experiment, FBP is applied only to the last Conv layer, which indirectly reduces the parameter count of the first FC layer (typically one of the largest) by decreasing its input channels.
Fig.~\ref{fig:lrf_fbp} shows the additional parameter and FLOPs reduction achieved by the combined method, normalized to the LRF-only baseline. This integration results in an extra 21\%–80\% reduction in memory usage.

%%%%%%%%%%%%%%%%%%%%%%%%%%%%%%%%%%%%%%%%%%%%%%%%%%%%%%%%%%%%%%
\begin{figure}[tbp]
\centerline{\includegraphics[trim={1mm 0 0 1mm}, clip,width=0.49\textwidth]{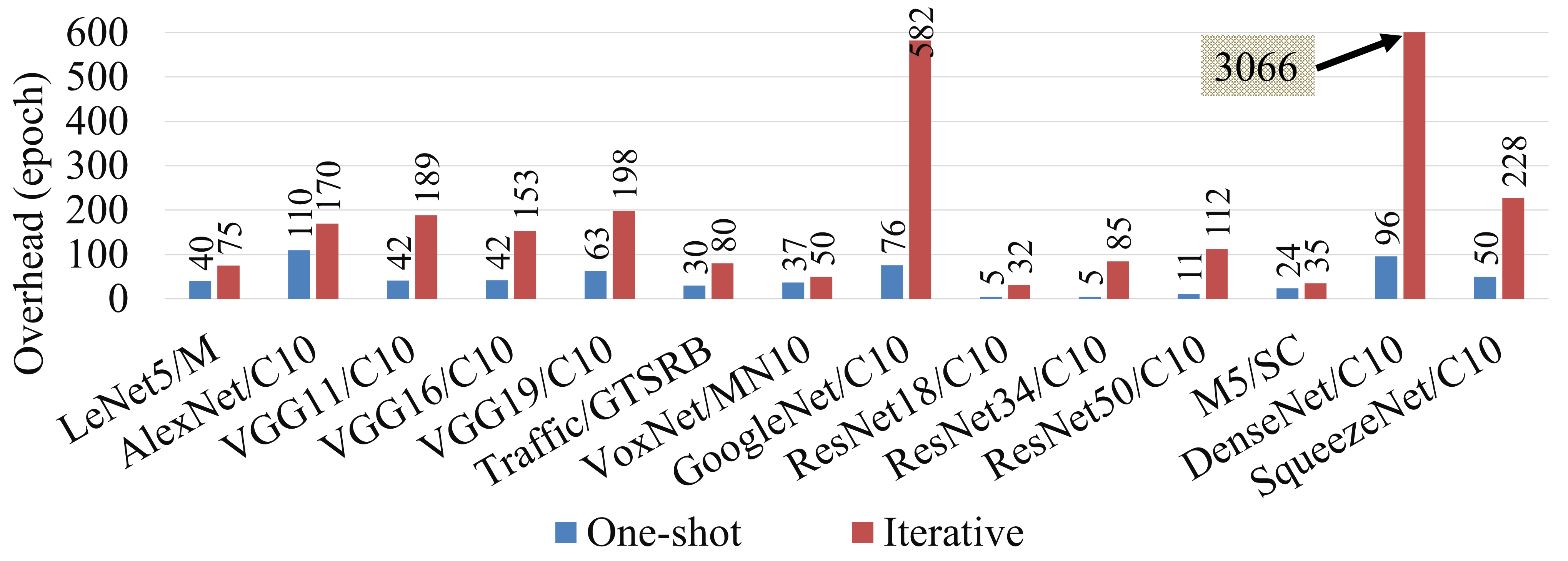}}

\squeezeup{}
\squeezeup{}
\squeezeup{}

\caption{Overhead of One-shot vs. iterative fine tuning.}

\squeezeup{}
\squeezeup{}
\squeezeup{}

\label{fig:tucker_vs_iterative}
\end{figure}

\noindent \textbf{Fine-Tuning Time Evaluation}
This Subsection evaluates the one-shot fine-tuning strategy used in our methodology. Unlike other approaches that re-train the model for each configuration~\cite{training1} or fine-tune after every layer is factorized~\cite{iteratively}, our method performs a single fine-tuning pass after all targeted layers are factorized, avoiding costly iterative calibration or retraining.
To assess the efficiency of this one-shot approach, we modified our framework to perform iterative fine-tuning, where each layer is factorized and immediately fine-tuned. For this experiment, we used Tucker decomposition with a step-size of 5 and max-sol of 3 for Conv layers, and SVD with a step-size of 2 for FC layers.

Fig.~\ref{fig:tucker_vs_iterative} compares the overhead time of the proposed method (one-shot) and iterative fine-tuning method. The overhead time accounts for the overall execution time required to construct the factorized model, such as decomposing the weights or calculating the similarity between the factorized and original models. This overhead time is normalized to the duration of a single training epoch for each model.
Only one dataset is shown for brevity.

Fig.~\ref{fig:tucker_vs_iterative} clearly demonstrates that the one-shot method consistently achieves significantly lower overhead in all evaluated scenarios. Specifically, the reduction in overhead reaches up to 31.9× (DenseNet model on C10 dataset), with an average reduction of approximately 8× across the entire models.
Furthermore, the gap in overhead becomes more pronounced as the complexity and size of the models increase. For larger and deeper models, characterized by more layers, the iterative fine-tuning process becomes increasingly computationally expensive due to repeated calibration over every single layer. In contrast, the one-shot method avoids this iterative burden by applying a targeted update in a single step, making it more scalable and efficient for modern, large-scale architectures. These results indicate that the proposed one-shot method becomes increasingly effective as model size and complexity grow, highlighting the suitability of the proposed method for efficiently handling the fine-tuning of large and deep models.

\section{Conclusion} \label{sec:conclusion}

\begin{table*}[]

\caption{Mapping of Quantitative Thresholds to Qualitative Levels Used in the Spider Graphs for Six Decomposition Methods Across Multiple Characteristics.}
\centering

\begin{tabular}{
>{\columncolor[HTML]{DAE8FC}}c ccccc}
\textbf{Metric} & \cellcolor[HTML]{DAE8FC}\textbf{5 = Excellent} & \cellcolor[HTML]{DAE8FC}\textbf{4 = Good} & \cellcolor[HTML]{DAE8FC}\textbf{3 = Moderate} & \cellcolor[HTML]{DAE8FC}\textbf{2 = Low} & \cellcolor[HTML]{DAE8FC}\textbf{1 = Poor} \\ \hline
\textbf{Rank   Configurations} & 5+ ranks & 4-5 ranks & 2–3 ranks & 1 rank + limits & Fixed \\ \hline
\textbf{Best   param count} & \textgreater{}98\% & 94-98\% & 90-94\% & 80-90\% & \textless{}80\% \\ \hline
\textbf{Worst   param count} & \textgreater{}20\% & 10-20\% & 6-10\% & 2-6\% & \textless{}2\% \\ \hline
\textbf{Best   FLOPs count} & \textgreater{}98\% & 94-98\% & 90-94\% & 80-90\% & \textless{}80\% \\ \hline
\textbf{Worst   FLOPs count} & \textgreater{}20\% & 10-20\% & 6-10\% & 2-6\% & \textless{}2\% \\ \hline
\textbf{Best   overall mem} & improve \textgreater 90\% & 60-90\% & 30-60\% & improve \textless 30\% & increase memory \\ \hline
\textbf{Worst   overall mem} & No extra memory & increase \textless 25\% & increase \textless 25-75\% & increase \textless 75-150\% & increase \textgreater 150\% \\ \hline
\textbf{ES} & Huge   \textgreater 10\textasciicircum{}6 & extra Large (10\textasciicircum{}4–10\textasciicircum{}6) & Large (10\textasciicircum{}3–10\textasciicircum{}4) & Medium (10²–10³) & Small (\textless{}100) \\ \hline
\textbf{Param coverage} & \textgreater{}98\% & 93-98\% & 85-93\% & 70-85\% & \textless{}70\% \\ \hline
\textbf{FLOPs   coverage} & \textgreater{}98\% & 93-98\% & 85-93\% & 70-85\% & \textless{}70\% \\ \hline
\textbf{Flexibility} & Shape + ranks & Per-dim ranks & Multiple ranks & Fixed rank & Rigid \\ \hline
\textbf{Decomposition   Time} & \textless{}5s & 5–30s & 30s-1min & 1min–5min & \textgreater{}5min \\ \hline
\end{tabular}
\label{table:metrics}

\end{table*}

This paper presents an end-to-end DSE methodology and framework that formulates LRF as a multi-objective optimization problem. Unlike existing approaches, our method introduces a more efficient rank selection strategy based on feature map similarity (rather than weight similarity), significantly reduces fine-tuning time, supports all CNN layer types and six LRF techniques, and is compatible with additional compression methods such as FBP. Notably, it also supports hybrid decomposition for the first time, allowing different LRF methods to be applied per layer to enhance overall compression.
Additionally, we present the first in-depth analysis of six LRF methods, uncovering several key insights. For example, we observe that LRF solutions are unevenly distributed across the Memory–FLOPs space. Moreover, for a fixed parameter count, different methods yield varying FLOPs and overall memory, depending on the layer shape (and vice versa). Furthermore, we highlight that no single LRF method is universally ideal, as each involves distinct trade-offs.

Our framework outperforms state-of-the-art techniques like VBMF and FBP in compression efficiency while preserving accuracy. Moreover, we demonstrate that LRF can be effectively combined with FBP for further gains. Future work will focus on refining similarity thresholds and extending the framework to other architectures, such as transformers.

\appendix

Table~\ref{table:metrics} shows the range of different metrics used in Fig.~\ref{fig:spider_conv} and Fig.~\ref{fig:spider_fc}. Higher score means better.
\textbf{Rank Configuration:}
This metric indicates the degree of configurability in selecting decomposition ranks for each method. It reflects how many independent ranks can be chosen.
\textbf{Best/Worst Param/FLOPs count:}
Denotes the maximum and minimum parameter/FLOPs reduction achieved by each method. The "best" value corresponds to the configuration yielding the highest parameter/FLOPs reduction, while the "worst" indicates the least parameter/FLOPs reduction within feasible configurations.
% \textbf{Best/Worst FLOPs count:}
% Denotes the maximum and minimum FLOPs reduction achieved by each method. The "best" value corresponds to the configuration yielding the highest FLOPs reduction (i.e., smallest number of FLOPs), while the "worst" indicates the least FLOPs reduction (or largest FLOPs count) within feasible configurations.
\textbf{Best/Worst overall mem:}
Measures the total memory usage, accounting for both parameters and intermediate feature maps.
The best case indicates whether overall memory usage improves (i.e., is reduced) compared to the original layer.
The worst case captures how much the total memory footprint increases, or at best, whether there is no increase.
\textbf{ES:}
Reflects the size of the practical design space for LRF solutions. This metric counts the number of distinct decomposition configurations that are feasible for a given method.
\textbf{Param/FLOPs coverage:}
Quantifies the variability or range in parameter and FLOPs reduction across different configurations. It is defined as the difference between the best and worst reduction values.
\textbf{Flexibility:}
Captures the method’s adaptability to different layer shapes and rank settings. Several flexibility modes are defined. \textit{Shape + Ranks:} The method supports diverse decomposition shapes for a single layer structure and allows setting multiple ranks. \textit{Per-dim ranks:} The method can selectively skip decomposing certain dimensions and assign independent ranks to others. \textit{Multiple ranks:} All tensor dimensions are decomposed, but each can have a distinct rank. \textit{Fixed:} A single rank must be applied uniformly across all dimensions. \textit{Rigid:} Applies only to matrix structures, with a single dimension and fixed rank.
\textbf{Decomposition Time:}
Indicates the computational cost (time) required to decompose a given weight tensor or matrix. This reflects the method’s practical overhead during model compression or deployment.

% \begin{itemize}
%     \item \textit{Shape + Ranks:} The method supports diverse decomposition shapes for a single layer structure and allows setting multiple ranks

%     \item \textit{Per-dim ranks:} The method can selectively skip decomposing certain dimensions and assign independent ranks to others

%     \item \textit{Multiple ranks:} All tensor dimensions are decomposed, but each can have a distinct rank

%     \item \textit{Fixed:} A single rank must be applied uniformly across all dimensions

%     \item \textit{Rigid:} Applies only to matrix structures, with a single dimension and fixed rank

% \end{itemize}

% \section*{Acknowledgment}

% The preferred spelling of the word ``acknowledgment'' in American English is without an ``e'' after the ``g.'' Use the singular heading even if you have many acknowledgments. Avoid expressions such as ``One of us (S.B.A.) would like to thank ... .'' Instead, write ``F. A. Author thanks ... .'' In most cases, sponsor and financial support acknowledgments are placed in the unnumbered footnote on the first page, not here.

% \bibliographystyle{IEEEtran}
\bibliographystyle{ieeetr}
% \bibliography{references_original}
% \bibliography{references_edited}
\bibliography{references_edited_full_info}

\end{document}